\documentclass[sigconf]{acmart}
\usepackage{colortbl}
\usepackage{multirow}
\usepackage{arydshln}
\usepackage{graphicx}
\usepackage{colortbl}
\usepackage{subcaption}  % 在导言区添加
\usepackage{xcolor}
\definecolor{myspringgreen}{RGB}{0,255,127}
\AtBeginDocument{%
  }

\setcopyright{acmlicensed}
\copyrightyear{2025}
\acmYear{2025}
\setcopyright{acmlicensed}\acmConference[MM '25]{Proceedings of the 33rd ACM International Conference on Multimedia}{October 27--31, 2025}{Dublin, Ireland}
\acmBooktitle{Proceedings of the 33rd ACM International Conference on Multimedia (MM '25), October 27--31, 2025, Dublin, Ireland}
\acmDOI{10.1145/3746027.3755405}
\acmISBN{979-8-4007-2035-2/2025/10}
\acmSubmissionID{3600}

\begin{document}

%%
%% The "title" command has an optional parameter,
%% allowing the author to define a "short title" to be used in page headers.
\title{Omni$^2$: Unifying Omnidirectional Image Generation and Editing in an Omni Model}

%%
%% The "author" command and its associated commands are used to define
%% the authors and their affiliations.
%% Of note is the shared affiliation of the first two authors, and the
%% "authornote" and "authornotemark" commands
%% used to denote shared contribution to the research.
% \author{Ben Trovato}
% \authornote{Both authors contributed equally to this research.}
% \email{trovato@corporation.com}
% \orcid{1234-5678-9012}
% \author{G.K.M. Tobin}
% \authornotemark[1]
% \email{webmaster@marysville-ohio.com}
% \affiliation{%
%   \institution{Institute for Clarity in Documentation}
%   \city{Dublin}
%   \state{Ohio}
%   \country{USA}
% }
% \newcommand{\corrauthmark}{\textsuperscript{\ref{corrauth}}}
\author{Liu Yang}
\affiliation{%
  \institution{SJTU}
  \city{Shanghai}
  \country{China}}
\email{ylyl.yl@sjtu.edu.cn}
\vspace{-1em}

\author{Huiyu Duan*}
\affiliation{%
  \institution{SJTU}
   \city{Shanghai}
  \country{China}}
\email{huiyuduan@sjtu.edu.cn}
\vspace{-1em}

\author{Yucheng Zhu}
\affiliation{%
  \institution{SJTU}
   \city{Shanghai}
  \country{China}}
\email{zyc420@sjtu.edu.cn}
\vspace{-1em}

\author{Xiaohong Liu}
\affiliation{%
  \institution{SJTU}
   \city{Shanghai}
  \country{China}}
\email{xiaohongliu@sjtu.edu.cn}

\author{Lu Liu}
\affiliation{%
  \institution{SJTU}
   \city{Shanghai}
  \country{China}}
\email{lettieliu@sjtu.edu.cn}

\author{Zitong Xu}
\affiliation{%
  \institution{SJTU}
   \city{Shanghai}
  \country{China}}
\email{xuzitong@sjtu.edu.cn}

\author{Guangji Ma}
\affiliation{%
  \institution{‌UESTC}
   \city{Shanghai}
  \country{China}}
\email{guangjima0806@gmail.com}

\author{Xiongkuo Min*}
\affiliation{%
  \institution{SJTU}
   \city{Shanghai}
  \country{China}}
\email{minxiongkuo@sjtu.edu.cn}
% \authornote{Corresponding authors.}

\author{Guangtao Zhai*}
\affiliation{%
  \institution{SJTU}
   \city{Shanghai}
  \country{China}}
\email{zhaiguangtao@sjtu.edu.cn}
% \authornote{Corresponding authors.}

\author{Patrick Le Callet}
\affiliation{%
  \institution{Université de Nantes}
   \city{Nantes}
   \country{France}}
\email{patrick.lecallet@univ-nantes.fr}
\thanks{*~Corresponding authors.}
\renewcommand{\shortauthors}{Liu Yang et al.}

%%
%% The abstract is a short summary of the work to be presented in the
%% article.
\begin{abstract}
  $360^{\circ}$ omnidirectional images (ODIs) have gained considerable attention recently, and are widely used in various virtual reality (VR) and augmented reality (AR)  applications. However, capturing such images is expensive and requires specialized equipment, making ODI synthesis increasingly important. While common 2D image generation and editing methods are rapidly advancing, these models struggle to deliver satisfactory results when generating or editing ODIs due to the unique format and broad 360$^{\circ}$ Field-of-View (FoV) of ODIs.
  To bridge this gap, we construct \textbf{\textit{Any2Omni}}, the first comprehensive ODI generation-editing dataset comprises 60,000+ training data covering diverse input conditions and up to 9 ODI generation and editing tasks.
  Built upon Any2Omni, we propose an \textbf{\underline{Omni}} model for \textbf{\underline{Omni}}-directional image generation and editing (\textbf{\textit{Omni$^2$}}), with the capability of handling various ODI generation and editing tasks under diverse input conditions using one model. Extensive experiments demonstrate the superiority and effectiveness of the proposed Omni$^2$ model for both the ODI generation and editing tasks.
  Both the Any2Omni dataset and the Omni$^2$ model are publicly available at: \textcolor{purple}{https://github.com/IntMeGroup/Omni2}.
\end{abstract}

%%
%% The code below is generated by the tool at http://dl.acm.org/ccs.cfm.
%% Please copy and paste the code instead of the example below.
%%
% \vspace{-2em}

%%
%% Keywords. The author(s) should pick words that accurately describe
%% the work being presented. Separate the keywords with commas.
%% A "teaser" image appears between the author and affiliation
%% information and the body of the document, and typically spans the
%% page.
\begin{teaserfigure}
\vspace{-1.5em}
  \includegraphics[width=\textwidth]{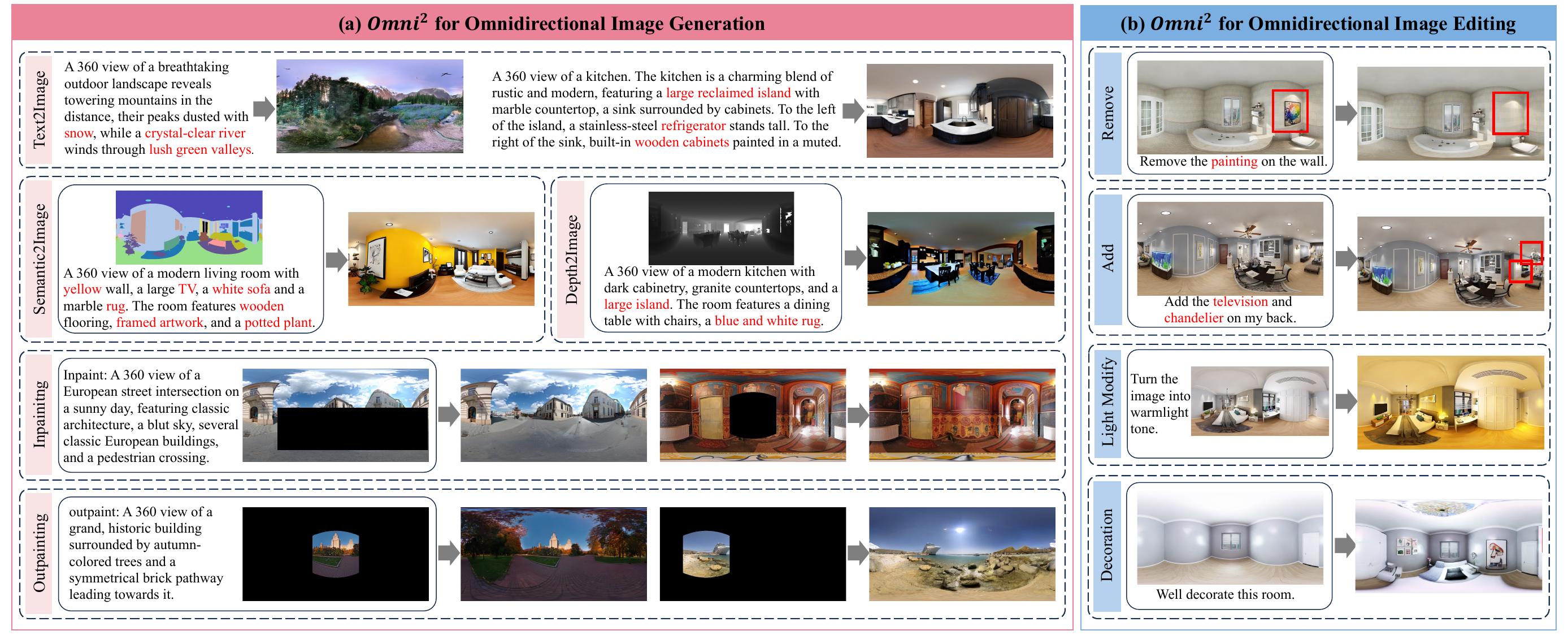}
  % \vspace{-20pt}
  \vspace{-2.5em}
  \caption{We propose the first omni model for omnidirectional image generation and editing, termed \textbf{\textit{Omni$^2$}}. \textit{Omni$^2$} is capable of handling both omnidirectional image generation and editing with various input conditions, demonstrating strong potential across diverse tasks, as demonstrated in (a) and (b).}
  \label{fig:teaser}
\end{teaserfigure}

% \received{20 February 2007}
% \received[revised]{12 March 2009}
% \received[accepted]{5 June 2009}

%%
%% This command processes the author and affiliation and title
%% information and builds the first part of the formatted document.
\maketitle

\begin{table*}[htbp]
  \centering
  \hspace*{0em} % 左移整个内容块
  \begin{minipage}[c]{0.34\textwidth} % 微调图片宽度
    \centering
    \includegraphics[width=\linewidth,height=0.7\linewidth,keepaspectratio]{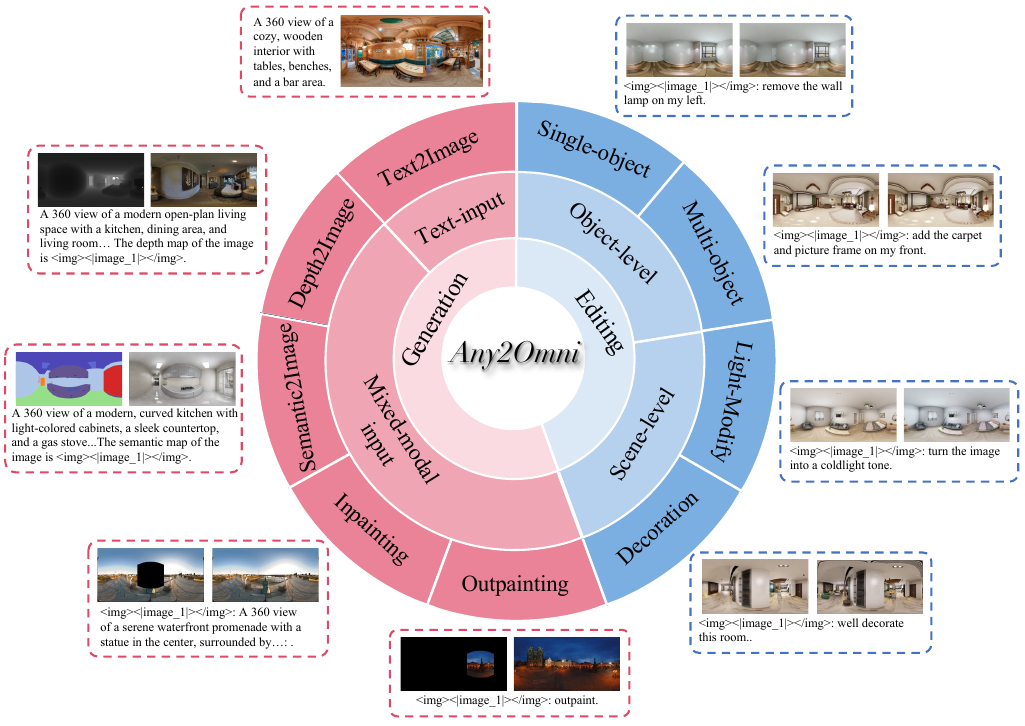}
  \end{minipage}%
  \hspace{0.01\textwidth} % 最小间距
  \begin{minipage}[c]{0.63\textwidth} % 微调表格宽度
    \centering
    \footnotesize
    \renewcommand{\arraystretch}{1.1} % 进一步压缩行高
    \setlength{\tabcolsep}{3.5pt} % 减小列间距
    \begin{tabular}{@{}lccccc@{}}
      \toprule
      Task Type & Category & Sub-category & Input Condition & \multicolumn{1}{c}{\# training} & \multicolumn{1}{c}{\# testing} \\
      \midrule
      \multirow{6}{*}{Generation} & \multirow{2}{*}{Text2Image} & short-caption & text & 10,000 & 1,000 \\
      & & detailed-caption & text & 10,000 & 1,000 \\
      & Semantic2Image & - & image+text & 6,500 & 500 \\
      & Depth2Image & - & image+text & 7,000 & 700 \\ 
      & Inpainting & - & image+text/image & 2,938/2,941 & 500/500 \\
      & Outpainting & - & image+text/image & 2,946/2,950 & 500/500 \\
      \hdashline
      \multirow{4}{*}{Editing} & \multirow{2}{*}{Object-level Editing} & single-object removal/add & image+text & 6,690 & 1000 \\
      & & multi-object removal/add & image+text & 5,889 & 500 \\
      & \multirow{2}{*}{Scene-level Editing} & Light-Modify & image+text & 4,698 & 500 \\
      & & Decoration & image+text & 5,187 & 500 \\
      \bottomrule
    \end{tabular}
    % \caption{Your table caption here}
  \end{minipage}
  \caption{Detailed data distribution in Any2Omni dataset.}
  \vspace{-2.5em}
  \label{tab1_dataset}
\end{table*} 
\vspace{-1em}
\section{Introduction}\label{intro} 
With the rapid advancement of virtual reality (VR) technology, 360$^{\circ}$ omnidirectional images (ODIs) have gained increasing attention. However, capturing ODIs requires expensive and specialized hardwares, making ODI synthesis a crucial task. While 2D image generation techniques have gradually matured with the rapid advancement of AIGC \cite{sd, chang2023muse, liu2024f, duan2025finevq, tian2024visual}, ODI generation remains underexplored. Previous works, such as MVDiffusion \cite{mvdiffusion}, can only generate ODIs with a limited vertical Field-of-View (FoV) of $360^{\circ}\times 90^{\circ}$, restricting its applicability in real-world scenarios. Though some other methods \cite{text2light, diffpano, panfusion} are capable of generating full $360^{\circ}\times 180^{\circ}$ ODIs, they primarily focus on text-driven ODI generation, while ignoring other input conditions such as narrow viewport images, semantic maps, depth maps, \textit{etc.}

High-quality 2D image editing datasets \cite{instructpix2pix, magicbrush, min2024perceptual, emuedit, xu2025lmm4edit} have recently driven the advancement of image editing models \cite{magicbrush, instructdiffusion}. In the context of ODIs, editing is of great importance for enhancing quality of experience in immersive environment. However, omnidirectional image editing remains unexplored. Unlike 2D images, ODIs are stored in the warped equirectangular projection (ERP) format, thus making common 2D image editing algorithms cannot be directly applied. An alternative approach is applying 2D editing models on the designated single view of an ODI after viewport splitting, but it is both time-consuming and inefficient, and may generate inconsistent views. Moreover, due to the unique depth and spatial characteristics of omnidirectional images, conventional 2D editing models struggle to understand the spatial relationships between viewports or even within individual ODI views, making the split-and-edit approach impractical. Therefore, dedicated ODI editing dataset and method are necessary to advance research in this domain.

Built upon the rapid progress in 2D image generation and editing, integrating these tasks into a unified framework has become increasingly popular \cite{omnigen, dreamomni,mige}. In this paper, we aim to unify ODI generation and editing into an efficient model. To this end, we first introduce \textbf{\textit{Any2Omni}}, the first comprehensive dataset for omnidirectional image generation and editing tasks. As shown in Table \ref{tab1_dataset}, our dataset integrates multiple input modalities for ODI generation tasks. For the newly defined omnidirectional image editing tasks, we start by proposing a simple yet effective pipeline capable of generating high-quality, object-level indoor editing samples. Additionally, we introduce two scene-level ODI editing tasks utilizing existing ODI datasets \cite{structured3d, yang2024aigcoiqa2024, yang2025quality, duan2022confusing}. Overall, our Any2Omni dataset comprises over \textbf{\textit{60,000}} training samples, covering \textbf{\textit{9}} categories of omnidirectional image generation and editing tasks with various input conditions.

Based on Any2Omni, we further introduce the first omni model for omnidirectional image generation and editing, termed \textbf{\textit{Omni$^2$}}. Omni$^2$ adopts a simple yet effetive Transformer-based framework to support $360^{\circ}\times 180^{\circ}$ high-quality omnidirectional image synthesis under a variety of multimodal input conditions. In contrast to existing diffusion-based ODI generation methods, which incorporate additional attention blocks for multi-view consistency \cite{mvdiffusion, panfusion}, we introduce a novel approach by executing viewport-based bidirectional attention within a unified Transformer. Our model demonstrates superior performance across a wide range of ODI generation and editing tasks with various input conditions, as shown in Fig. \ref{fig:teaser}.

The main highlights of this work include:
\vspace{-3pt}
\begin{itemize}
\item We unify the ODI generation and editing tasks. To the best of our knowledge, our study is the first work to achieve multimodal input-based ODI generation for various tasks, and the first to explore ODI editing.
\item We construct \textbf{Any2Omni}, the first comprehensive dataset for omnidirectional image generation and editing that contains over \textit{60,000} training samples covering \textit{9} ODI generation and editing tasks with various input conditions.
\item We propose \textbf{Omni$^2$}, the first omni model for omnidirectional image generation and editing. Omni$^2$ is capable of processing various input conditions and producing high-quality, viewport-consistent ODIs.
\item Extensive experimental results demonstrate that the proposed Omni$^2$ model exhibits state-of-the art performance on ODI generation tasks, and shows great potential on ODI editing tasks.
\end{itemize}
\vspace{-9pt}
\section{Related Work}
\subsection{Omnidirectional Image Generation}
Although 2D image generation has made significant progress recently, omnidirectional image generation remains notably limited. Prior works, such as MultiDiffusion \cite{multidiffusion}, employ pretrained diffusion model to generate long images from text input. However, these images are not composed of stitched multi-view images thus are not aligned with the true projection process of ODIs. MVDiffusion \cite{mvdiffusion} solves this problem by synchronously generating eight overlapping views using pretrained stable diffusion model \cite{sd} and introduces Correspondence-Aware Attention (CAA) module to ensure viewport consistency. However, it only generates the central perspectives, omitting the top and bottom views, limiting its applicability in real-world scenarios.
Other methods like DiffPano \cite{diffpano} and PanFusion \cite{panfusion} are able to generate omnidirectional images with the full FoV, yet they are still limited to specific input conditions, and lack support for generation tasks with other input conditions, such as depth-to-image generation and outpainting, \textit{etc}.
% To bridge this gap, we introduce Any2Omni, a dataset encompassing a wide range of ODI generation tasks with various types of condition input, as illustrated in Fig. \ref{fig:teaser} (b). Furthermore, we propose a unified ODI generation model that achieves superior performance across various ODI generation tasks.
\vspace{-6pt}
\subsection{Independent Image Generation and Editing}
Recent advancements in 2D generative models have been remarkable, with methods such as the Stable Diffusion series \cite{sd, sdxl} and DALL-E \cite{ramesh2022hierarchical} demonstrating superior generative capabilities. In addition, the emergence of large-scale, high-quality 2D image editing datasets \cite{instructpix2pix, magicbrush, emuedit, instinpaint} and benchmarks \cite{imagen} have significantly advanced image editing models \cite{emuedit, smartedit, omniedit}. 
However, limited works have been proposed for ODI generation as aforementioned, and the field of ODI editing remains unexplored, despite the essential role ODI plays in real-world applications.
\vspace{-12pt}
\subsection{Unified Image Generation and Editing}
With the rapid advancement of 2D image generation and editing models, unified 2D generation-editing datasets and models have emerged. Compared to single-task models such as text-to-image (T2I) generation or image inpainting, unified multimodal-input generation and editing models offer broader applications. DreamOmni \cite{dreamomni} introduces a synthetic data pipeline for generating large-scale multi-task image generation and editing training examples and constructs a unified image generation and editing model. OmniGen \cite{omnigen} constructs a large-scale, diverse dataset, X2I, encompassing various image generation and editing tasks for joint training, and further proposes a unified diffusion-transformer model OmniGen, which demonstrates strong capabilities across various text-to-image generation and downstream tasks.
However, in the field of ODI synthesis, there are still many unexplored tasks, let along comprehensive datasets and models.
To address these issues and to advance research on omnidirectional image synthesis, we introduce Any2Omni, the first comprehensive dataset for ODI generation and editing, and Omni$^2$, a unified model that shows strong capabilities across multiple ODI generation and editing tasks.
\vspace{-6pt}
\section{Any2Omni Dataset}\label{sec:pipeline}
%\begin{figure*}[t]\centering
%\vspace{-0.8em}
%\includegraphics[width=0.98\linewidth]{figs/figs/fig1.pdf}
%\vspace{-0.6em}
%\caption{
%}
%\label{fig:1_frontpage}
%\vspace{-1em}
%\end{figure*}
\begin{figure*}[t]\centering
\vspace{-1em}
\includegraphics[width=0.95\linewidth]{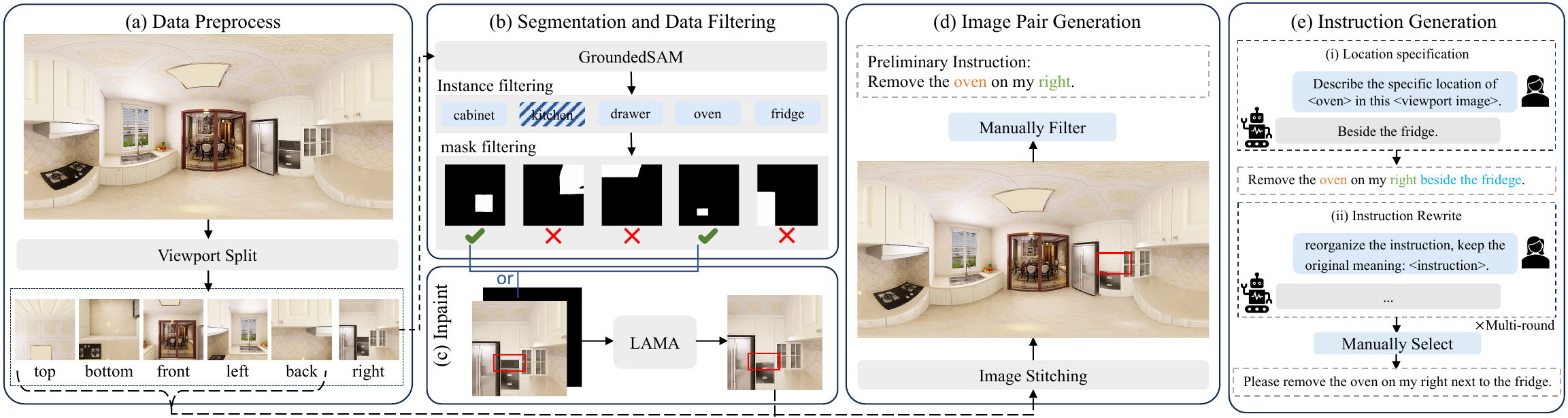}
\vspace{0em}
\captionof{figure}{We present a simple yet effective pipeline for constructing high-quality object-level indoor omnidirectional image editing dataset. Our pipeline is mainly consisted of two parts, \textit{i.e.}, image pair generation and instruction refinement. (a) The ODI input undergoes viewport splitting to generate six perspective images.
(b) Viewport image is processed through a segmentation model for instance-level segmentation and class labeling, followed by dual-stage filtering for quality control.
(c) A selected instance is removed via inpainting \cite{lama}. (d) The edited viewport image is seamlessly stitched with other perspective images to form edited ODI.
(e) InternVL2-5 \cite{internvl} is deployed to refine editing instructions, adding positional details and enhancing linguistic diversity.
}
\label{fig:pipeline}
\vspace{-1.5em}
\end{figure*}
\vspace{0pt}

To gain the capability of handling various image generation and editing tasks, it is essential to build a comprehensive dataset for model training. Though some omnidirectional image generation works have tried training on captioned existing ODI datasets \cite{matterport3d, sun360, lavalindoor, panodiff}, a truly comprehensive ODI generation dataset remains absent. Furthermore, there's currently no dataset for omnidirectional image editing. 
In this section, we introduce \textbf{\textit{Any2Omni}}, the first comprehensive dataset for omnidirectional image generation and editing. For generation subset, we construct a diverse ODI generation dataset to support multiple ODI generation tasks with various input conditions. For ODI editing subset, we start by proposing a simple yet effective pipeline for object-level indoor ODI editing, and further propose a range of insightful indoor ODI editing tasks. Our \textbf{\textit{Any2Omni}} contains over \textbf{\textit{60,000}} training examples with \textbf{\textit{9}} tasks, establishing it as the first large-scale and comprehensive dataset dedicated to advancing research in \textbf{\textit{ODI generation and editing}}.
\vspace{-6pt}
\subsection{Generation Subset}
Although prior works have explored T2I and outpainting tasks for ODIs \cite{mvdiffusion, panfusion, SIG-SS, panodiff, wang2024360dvd}, they address these tasks separately, each tailored to a specific input-output format. Besides, we believe there are insightful ODI generation tasks that remain unexplored. To this end, we construct a comprehensive generation subset, encompassing 5 ODI generation tasks: text2ODI, inpainting, outpainting, semantic2ODI, and depth2ODI, among which we newly proposed semantic2ODI and depth2ODI as we believe they present significant opportunities for advancing ODI generation. We collect ODIs from existing datasets, including SUN360 \cite{sun360}, Structured3D \cite{structured3d}, \textit{etc., } and caption these images using InternVL2-5 \cite{internvl} to generate text condition. For inpainting and outpainting tasks, images are randomly masked as input image conditions. For semantic2ODI and depth2ODI tasks, we use paired images from Structured3D and Pano3D \cite{pano3d}, respectively, and generate captions to provide text conditions for these tasks. The detailed construction process is provided in the supplementary materials.
\vspace{-6pt}
\subsection{Editing Subset}
While 2D image editing has achieved significant advancement, the field of omnidirectional image editing remains unexplored, primarily due to the scarcity of high-quality training datasets. As object removal and addition are among the most fundamental tasks in image editing, in this work, we begin to address the challenge of constructing ODI editing dataset with a specific focus on the object-level removal and addition tasks. Additionally, we propose two scene-level ODI editing tasks that effectively leverage the existing ODI dataset \cite{structured3d}, contributing to the advancement of ODI editing.
\vspace{-3pt}
\subsubsection{Object-level Indoor Editing Pipeline}
The primary challenge in generating ODI editing datasets lies in creating editing image pairs where only a partial set of objects is precisely added or removed, while the other details of the image are retained.
Due to the rarity of ODIs in social media applications, it is not feasible to directly collect editing image pairs through web crawling, as is done with 2D editing datasets \cite{instructdiffusion, realedit}.
Since constructing manually annotated datasets is impractically time-consuming, 2D image editing datasets are often built by either applying 2D generation methods \cite{instructpix2pix} or utilizing existing 2D-related datasets. For instance, Inst-Inpaint \cite{instinpaint} leverages existing dataset \cite{gqa} to obtain scene graphs within the image, which significantly simplifies the task.
% Instead, we can only rely on synthetic data generation methods.
% Directly applying 2D editing models on omnidirectional image or a splited viewport is unrealistic as described in section \ref{intro}.
% Since constructing manually annotated datasets is impractically time-consuming, some works \cite{instructpix2pix} utilizes the Prompt-to-Prompt (P2P) approach \cite{prompt2prompt}, which relies on word-to-word alignment between the input image caption and the edited image caption to generate editing image pairs. However, P2P is highly dependent on the alignment of input and output captions, and the current generation approach for omnidirectional images is not as mature as that for 2D images, making it challenging to ensure the quality of generated images. Other works \cite{instinpaint} leverage existing datasets \cite{gqa} to obtain scene graphs within the image, which significantly simplifies the task.
However, due to the immaturity of current ODI generation methods and limited sources of ODI-related datasets, we are forced to build the omnidirectional image editing dataset from scratch. In response, we develop a pipeline that can generate high-quality editing image pairs and detailed editing instructions from a single, unconditioned indoor ODI input, as illustrated in Fig. \ref{fig:pipeline}.

Compared to outdoor omnidirectional images, indoor ODIs are often more diverse and contain easily segmentable objects. As an initial attempt at omnidirectional image editing, we select Structured3D \cite{structured3d}, a large-scale indoor ODI dataset, to generate object-level removal/addition image pairs. Inspired by \cite{instinpaint}, we constructed our editing image pairs through a pipeline that integrates segmentation and inpainting models to generate precise edited image pairs. However, unlike Inst-Inpaint, which relies on object categories provided by the scene graphs in the GQA dataset \cite{gqa} for instance segmentation, our task requires an algorithm capable of both detecting instances and outputting corresponding instance categories. To accomplish this, we choose GroundedSAM \cite{groundedsam} due to its ability to effectively detect and classify instances in a single pass.
Directly applying segmentation on the ERP formatted ODIs yields imprecise results. Furthermore, given the richness of objects in indoor scenes, we propose to generate \textbf{editing pairs with predefined viewports}. Specifically, we first split the ODIs into six viewpoints: front, back, left, right, top, and bottom. Then, segmentation is applied on each individual viewport, as shown in Fig. \ref{fig:pipeline}(a).
% With the selected instance classes and segmentation masks predicted by GroundedSAM, we adopt LAMA \cite{lama} to inpaint them and thus generating object-removal outputs, shown in Fig. \ref{fig:pipeline} (c)-(e). Inspired by \cite{instructdiffusion, instinpaint}, object-adding pairs are generated by swapping the input and output images.

We perform dual-stage post-processing after the segmentation process to ensure data quality, as illustrated in Fig. \ref{fig:pipeline}(b). First, we apply instance filtering to remove instances that are irrelevant for our object-level editing purpose, such as ``floor'', ``cityscape'', and ``bedroom''.
Since the inpainting operation is applied to each individual viewport, instances that span across multiple viewports may experience disruptions in scene consistency when inpainting is applied to each viewport separately. To address this, we implement an automatic instance filtering method, where instances with segmentation masks located at the image edges are discarded. This process effectively filters out approximately $60\%$ of the segmentation results.

With the selected instance classes and corresponding segmentation masks, we adopt LAMA \cite{lama} to inpaint the instances, thereby generating object-removal outputs, shown in Fig. \ref{fig:pipeline}(c)-(d). The outputs are manually selected to ensure image quality. Inspired by \cite{instructdiffusion, instinpaint}, object-addition pairs are generated by swapping the input and output images.
By executing multiple rounds of the pipeline on the same perspective image and stitching together images edited from various viewports, we further obtain a set of \textbf{multi-object editing image pairs}.

% Editing instructions play a critical role in the editing process, requiring both diversity and precision to enhance model performance.
Preliminary editing intructions are formed using the detected instance class and the viewport information in the format as ``Remove the <instance class> on my <viewport>.''
In certain editing cases, the relative location of the object should be pointed out. To address this, we input both the original image and segmented instance class into Internvl2-5 \cite{internvl}, instructing it to generate relative position descriptions, as shown in Fig. \ref{fig:pipeline}(e). With these relative position descriptions, we create more detailed editing instructions: ``Remove the <instance class> on my <viewport> beside the <reference instance>.'' For multi-object editing, the instructions are formed as: ``Add the <instance class1> beside the <reference instance1> on my my <viewport1> and <instance class2> on my <viewport2>.''
To ensure data diversity, we further utilize Internvl2-5 \cite{internvl} to refine the generated editing instructions, thereby creating a range of diverse expressions that contribute to improving the model's robustness. The instruction generation process is carried out over multiple iterations, during which we manually select the most appropriate instructions.
%\begin{figure*}[t]\centering
%\vspace{-0.8em}
%\includegraphics[width=0.98\linewidth]{figs/figs/fig1.pdf}
%\vspace{-0.6em}
%\caption{
%}
%\label{fig:1_frontpage}
%\vspace{-1em}
%\end{figure*}

%\vspace{-1em}
\begin{figure*}[t]\centering
\vspace{-1.3pt}
\includegraphics[width=0.9\linewidth]{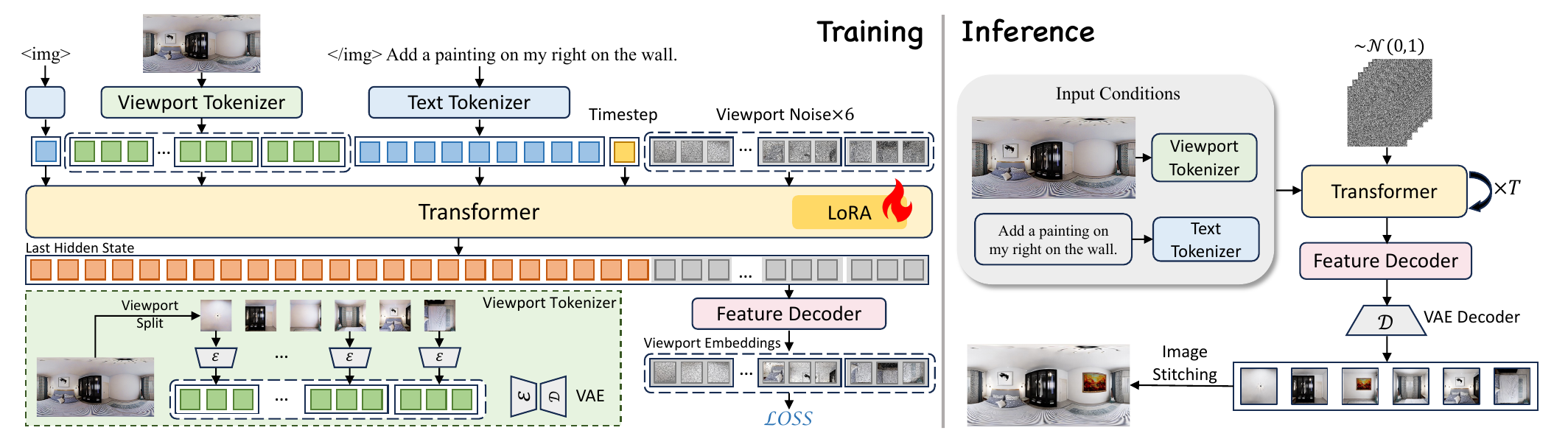}
\vspace{-1em}
\captionof{figure}{Overview of proposed Omni$^2$. Input prompt and Image are tokenized through text tokenizer and viewport tokenizer seperately before feeding into a simple yet effective transformer for viewport image generation. During inference, the generated viewports are seamlessly integrated to reconstruct a high-quality omnidirectional image.
}
\label{fig:model}
\vspace{-1.5em}
\end{figure*}
% \vspace{-12pt}
\vspace{-9pt}
\subsubsection{Scene-level ODI Editing}
The ability to manipulate entire scenes is a defining characteristic that enhances immersion and interactivity in omnidirectional images within virtual reality applications. To this end, we introduce two novel scene-level indoor editing tasks, namely light modify and indoor decoration, designed to enable comprehensive editing of ODIs.
We collect the image pairs from Structured3D and standardize the tasks into the input-output pair format, as detailed in the supplementary.

\vspace{-9pt}
\section{Proposed Method}
In this section, we introduce our \textbf{\textit{unified}} model, \textbf{Omni$^2$}, towards unifying omnidirectional image generation and editing from various condition inputs in free form using one model.
\vspace{-6pt}
\subsection{Overall Architecture}
The overall Architecture of Omni$^2$ is presented in Fig. \ref{fig:model}, which adopts a pretrained Transformer as the denoising network. The model takes arbitrarily interleaved text and images in free form as input conditions, which are then tokenized and fed into the Transformer for denoising, along with timestep tokens and viewport noise. The last hidden state is then passed through a feature decoder to obtain the final viewport embeddings. During inference, the viewport embeddings are subsequently fed into a VAE decoder to generate viewport images, which are then stitched together to form a seamless ODI output.
\vspace{-6pt}
\subsection{Model Design}
\subsubsection{Input Embedding}
 We propose generating separate viewports instead of treating the entire ODI as a single entity, in order to ensure viewport consistency and align with the capture process of ODIs. To address this, the noise input is defined on a viewport basis for the viewport-based diffusion process, as shown in Fig. \ref{fig:model}.
 
 Input image condition is also viewport tokenized to align with the output. Given an ODI as an input image, we employ a viewport tokenizer to convert the image into viewport-based tokens.
 Specifically, the input ODI is divided into six overlapping viewports, each of which is then transformed into viewport-based latent representations using a frozen VAE encoder. These representations are then flattened into a sequence of visual tokens with the patch size set to 2 following \cite{peebles2023scalable}. During training, the viewport noise is formulated as:
 \begin{align}
 \vspace{-3pt}
     z^i_t = tz^i+(1-t)\epsilon ^i
     \label{eq:1}
 \vspace{-3pt}
 \end{align}
 where $z^i$ is the viewport-based latent representation of the ground truth $\{x_i\}_{i=1}^{6}$, $t$ is the diffusion timestep and $\epsilon ^i \sim \mathcal{N}(0,I)$ is the Gaussian noise. During the inference stage, each viewport noise is formulated as a random sampled Gaussian noise, as shown in Fig. \ref{fig:model}.
The text prompt and timestep are also tokenized and fed into the pretrained transformer to facilitate the denoising process.
\subsubsection{Attention Mechanism for Viewport Consistency}
In omnidirectional image generation, maintaining viewport consistency is especially crucial.
% Unlike previous works that introduce additional trainable attention modules to address this issue \cite{mvdiffusion}, we leverage and modify the attention mechanism within the Transformer architecture itself to achieve multi-view attention.
Previous works often adopt diffusion model for ODI generation, with external attention modules designed to ensure viewport consistency \cite{mvdiffusion, panfusion}. However, these approaches introduce additional training parameters and lack holistic control across viewports, as viewports are generated seperately \cite{mvdiffusion}. In contrast, we propose to leverage and modify the attention mechanism within the Transformer architecture itself to achieve multi-view consistency.
Specifically, we adopt the Transformer from Phi-3 \cite{phi-3} as the denoising network. We believe that viewports should be treated equally during the generation process to maintain viewport consistency. To this end, we propose a novel approach to apply bidirectional attention mechanism within the viewport sequences while maintain the causal attention machanism elsewhere. This approach is not only simpler but also effectively preserves viewport consistency. Furthermore, since viewports are generated using the shared model, it allows for greater overall control over the image, ensuring more coherent results across all generated views.
%\begin{figure*}[t]\centering
%\vspace{-0.8em}
%\includegraphics[width=0.98\linewidth]{figs/figs/fig1.pdf}
%\vspace{-0.6em}
%\caption{
%}
%\label{fig:1_frontpage}
%\vspace{-1em}
%\end{figure*}

%\vspace{-1em}
\begin{figure*}[t]\centering
\vspace{-1em}
\includegraphics[width=0.85\linewidth]{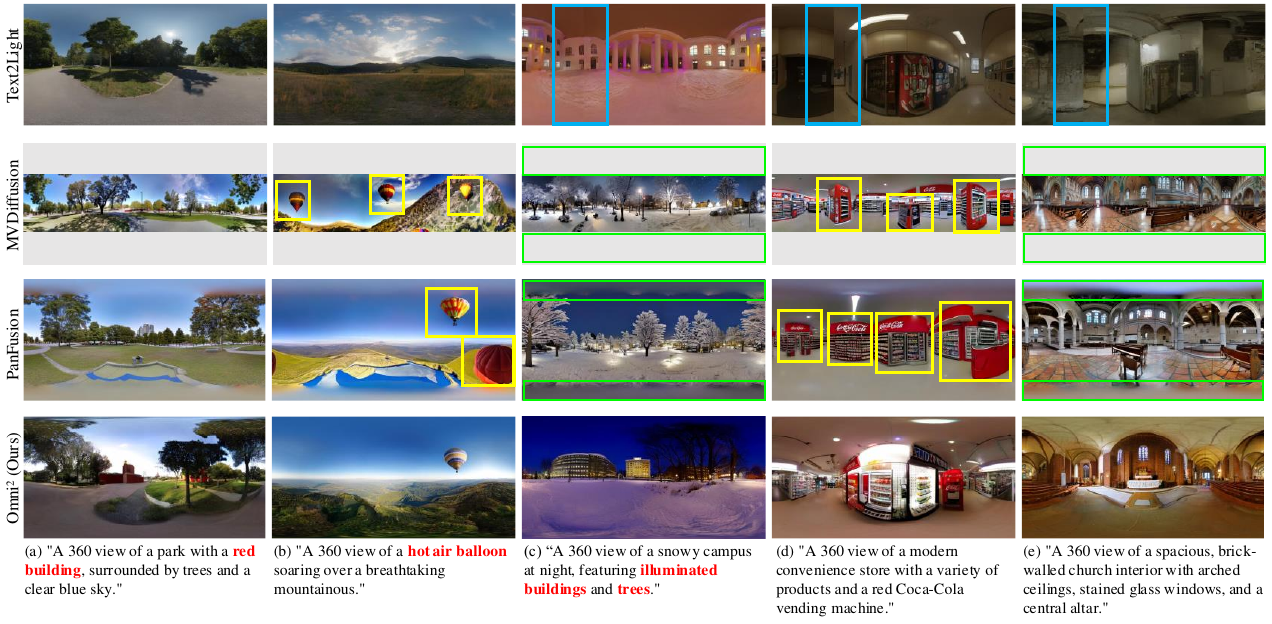}
\vspace{-1.5em}
\captionof{figure}{Text to ODI comparison between Text2Light \cite{text2light}, MVDiffusion \cite{mvdiffusion}, PanFusion \cite{panfusion} and ours. We highlight the \textcolor{cyan}{left-right inconsistency}, \textcolor{yellow}{repeating objects in different views} and \textcolor{myspringgreen}{top-bottom blurriness/missing} with corresponding color boxes. Objects that are missing in some baselines but present in our method are \textbf{\textcolor{red}{bolded and highlghted in red}} in the prompts.
}
\label{fig:comparison_generation}
\vspace{-1.5em}
\end{figure*}
% \vspace{-6pt}
\vspace{-6pt}
\subsubsection{Omnidirectional Image Generation}
The final hidden state from the Transformer is passed into a feature decoder to map the language space features into latent representations for each viewport. We utilize the final layer from \cite{peebles2023scalable} to accomplish this process. The VAE decoders then decode the generated viewport representations into predicted viewports, which are subsequently stitched together to form a seamless omnidirectional image.
\vspace{-6pt}
\subsection{Training}
\subsubsection{LoRA-based Finetuning}
2D generation models possess excellent prior knowledge of image generation and strong text comprehension capabilities. Due to the limited scale of omnidirectional images, we use the VAE from SDXL \cite{sdxl} and initialze the Transformer with pretrained weights from \cite{omnigen} to take advantage of the excellent generation ability on 2D images.
During training, only the Transformer is finetuned using LoRA \cite{lora} to improve training efficiency and retain generalization capability as much as possible.
\vspace{-6pt}
\subsubsection{Loss Function}
Flow matching \cite{flowmatching} is employed to train the model. We extend the loss from single-view to multi-view. For each training step, we randomly sample a timestep $t$ for all the viewport images $\{x_i\}_{i=1}^N$. The loss function is defined as:
% \vspace{0pt}
\begin{align}
    \mathcal{L} = \mathbb{E}[\sum_{i=1}^{N}||(z^i-\epsilon ^i)-v_{\theta}(z^i_t,t,c)||^2]
\label{eq:2}
\end{align}
% \vspace{-3pt}
% \{ z^i = \mathcal{E}(x_i)\}_{i=1}^N,\{ \epsilon ^i \sim \mathcal{N}(0,1)\}_{i=1}^N
where $z^i = \{\mathcal{E}(x_i)\}_{i=1}^N$ represents the latent embeddings of the $i$-th viewport image,  $\epsilon ^i \sim \mathcal{N}(0,1)\}_{i=1}^N$ is the Gaussian noise, $z^i_t$ is the noised latent for the $i$-th viewport as defined in Eq. \ref{eq:1}, and $c$ denotes the embeded condition.

For tasks like light modify and object-level editing, different regions should be adjusted with varying intensities. Specifically, for light modify, distinct areas require different levels of adjustment, while in object-level editing, only a small portion of the regions needs to be modified, thus we make slight modifications to the loss function for these tasks to enable the model to learn the appropriate intensities and to prevent it from simply copying the input image as the output, as discussed in \cite{omnigen}. This was achieved by introducing a weighted loss, where regions that differ significantly from the input image are assigned higher weights. However, although the decoration task belongs to editing, since the change is made to the overall image, applying weighted loss leads to unsatisfatory results. Therefore, we use the original loss for the decoration task.
\vspace{-6pt}
\section{Experiment}
\begin{table}

\centering

\caption{Comparison with state-of-the-art methods on text to omnidirectional image task. *Since MVDiffusion cannot generate full-FOV ODIs, we evaluate its performance here for reference purposes.
}
\vspace{-1.3em} 
\centering
\resizebox{\linewidth}{!}{
\begin{tabular}{lccccc}
   \toprule
   Methods& FAED$\downarrow$& FID$\downarrow$&IS$\uparrow$&CS$\uparrow$&Inference Time(s)$\downarrow$\\
     \midrule
   Text2Light~\cite{text2light}&2.70&91.05&4.90&0.7007&135.36 \\
   {MVDiffusion*} \cite{mvdiffusion}&3.06&92.59&6.64&0.5367&252.08\\
   PanFusion~\cite{panfusion}&2.54&80.66&{7.36}&0.8463&67.83 \\
   SD+LoRA~\cite{sd}&2.30&57.97&7.41&0.8540&31.25\\
   \rowcolor{gray!20}
   Omni$^2$ (Ours)&\textbf{2.25}&\textbf{47.32}&\textbf{7.62}&\textbf{0.8887}&\textbf{22.55}\\
   \bottomrule
  \end{tabular}}
  \vspace{-1.5em}
  \label{tab:generation}
\end{table}
  % \vspace{-1.5em}

\begin{table}

\centering

\caption{User study of text to ODIs.}
\vspace{-1.3em} 
\centering
\resizebox{\linewidth}{!}{
\begin{tabular}{lccc}
   \toprule
   Methods&Image Quality$\uparrow$&Image-Text Consistency$\uparrow$&Omni-Scene Consistency$\uparrow$\\
     \midrule
   Text2Light~\cite{text2light}&2.17&2.28&2.78\\
   PanFusion~\cite{panfusion}&3.33&3.94&4.44\\
   \rowcolor{gray!20}
   Omni$^2$ (Ours)&\textbf{4.06}&\textbf{4.72}&\textbf{4.83}\\
   \bottomrule
  \end{tabular}}
  \vspace{-2em}
  \label{tab:userstudy}
\end{table}
  % \vspace{-1.5em}

\subsection{Implementation Details}
\subsubsection{Training Settings}
%\begin{figure*}[t]\centering
%\vspace{-0.8em}
%\includegraphics[width=0.98\linewidth]{figs/figs/fig1.pdf}
%\vspace{-0.6em}
%\caption{
%}
%\label{fig:1_frontpage}
%\vspace{-1em}
%\end{figure*}

%\vspace{-1em}
\begin{figure*}[t]\centering
\vspace{-0.8em}
\includegraphics[width=1\linewidth]{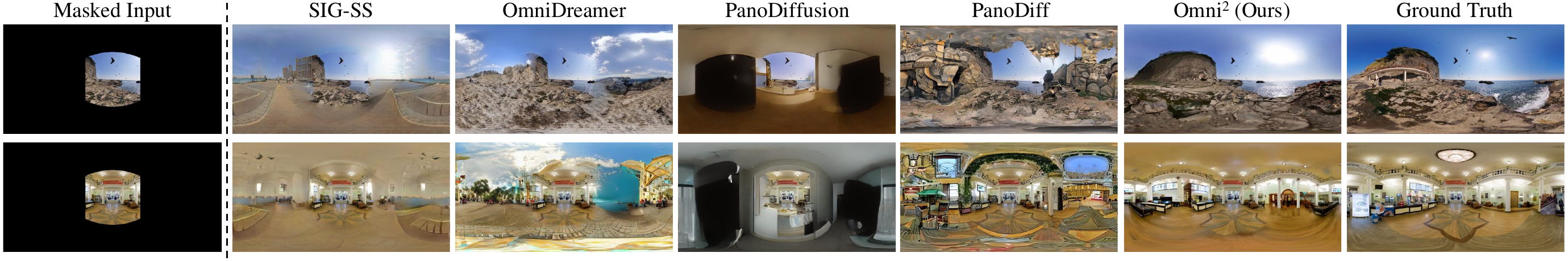}
\vspace{-2.5em}
\captionof{figure}{Outpainting comparison between SIG-SS \cite{SIG-SS}, OmniDreamer \cite{omnidreamer}, PanoDiffusion \cite{PanoDiffusion}, PanoDiff \cite{panodiff}, and ours.
}
\label{fig:comparison_outpaint}
\vspace{-1em}
\end{figure*}
% \vspace{-6pt}
%\begin{figure*}[t]\centering
%\vspace{-0.8em}
%\includegraphics[width=0.98\linewidth]{figs/figs/fig1.pdf}
%\vspace{-0.6em}
%\caption{
%}
%\label{fig:1_frontpage}
%\vspace{-1em}
%\end{figure*}

%\vspace{-1em}
\begin{figure}[t]\centering
\vspace{0em}
\includegraphics[width=1\linewidth]{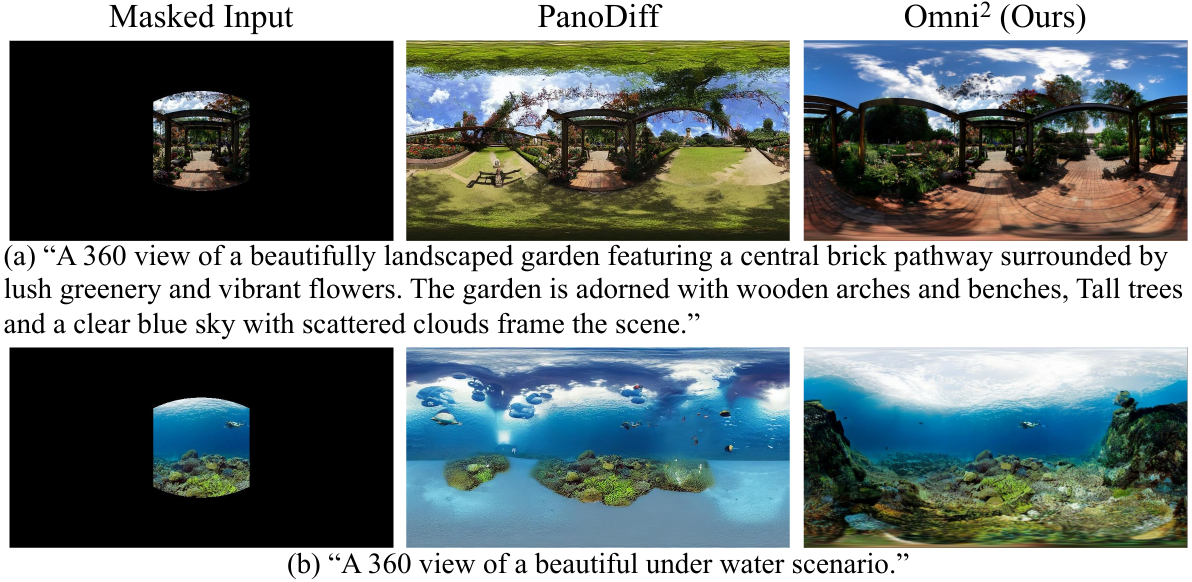}
\vspace{-2.5em}
\captionof{figure}{Text-instructed outpainting comparison between PanoDiff \cite{panodiff} and ours.
}
\label{fig:comparison_outpaint1}
\vspace{-1.5em}
\end{figure}
% \vspace{-6pt}
%\begin{figure*}[t]\centering
%\vspace{-0.8em}
%\includegraphics[width=0.98\linewidth]{figs/figs/fig1.pdf}
%\vspace{-0.6em}
%\caption{
%}
%\label{fig:1_frontpage}
%\vspace{-1em}
%\end{figure*}

%\vspace{-1em}
\begin{figure*}[t]\centering
\vspace{-1em}
\includegraphics[width=0.95\linewidth]{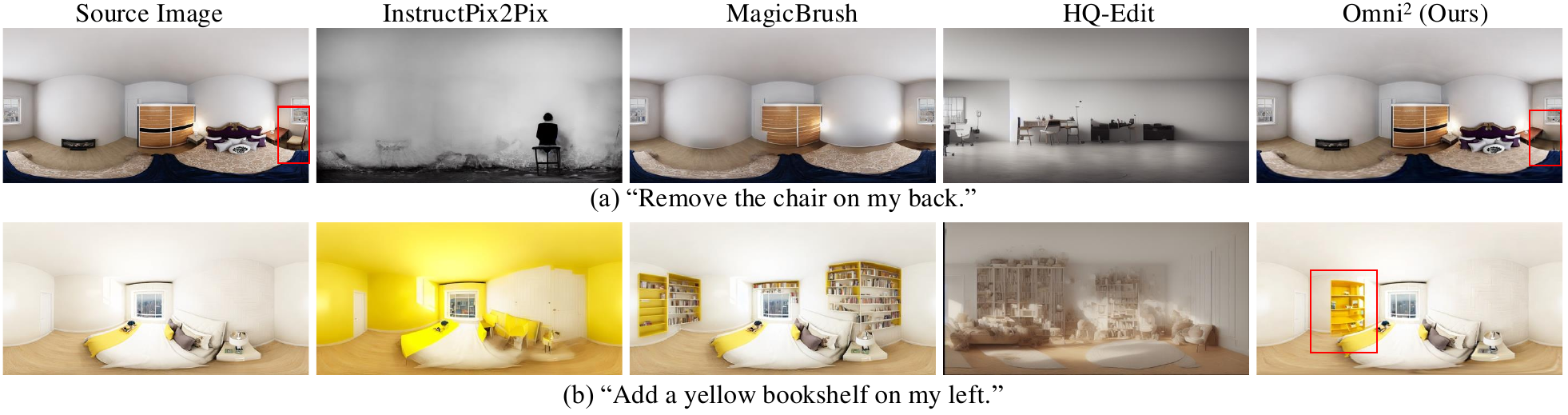}
\vspace{-1.5em}
\captionof{figure}{Visual comparison of directly applying existing 2D editing methods \cite{instructpix2pix, magicbrush, hqedit} on ODIs versus our method.
}
\label{fig:comparison_editing}
\vspace{-1.5em}
\end{figure*}
% \vspace{-6pt}
%\begin{figure*}[t]\centering
%\vspace{-0.8em}
%\includegraphics[width=0.98\linewidth]{figs/figs/fig1.pdf}
%\vspace{-0.6em}
%\caption{
%}
%\label{fig:1_frontpage}
%\vspace{-1em}
%\end{figure*}

%\vspace{-1em}
\begin{figure}[t]\centering
% \vspace{-1em}
\includegraphics[width=0.95\linewidth]{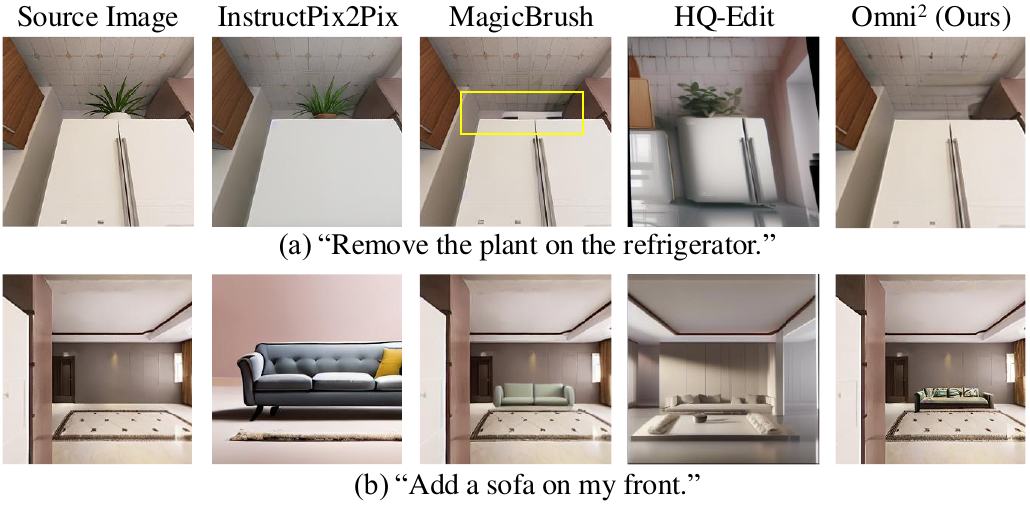}
\vspace{-1.5em}
\captionof{figure}{Visual comparison of directly applying existing 2D editing methods \cite{instructpix2pix, magicbrush, hqedit} on designated views of ODIs versus our method. Only the designated view is presented.
}
\label{fig:comparison_editing1}
\vspace{-2em}
\end{figure}
% \vspace{-6pt}
We train the Omni$^2$ model on our Any2Omni dataset using an \textbf{all in one} training strategy, \textit{i.e.}, all nine tasks, including both generation and editing, are trained simultaneously. The image resolution is set to $512\times 1024$ and the viewport FoV is set to $110^{\circ}$ with resolution of $256\times 256$. The total number of training step is 10k. The training is performed using the AdamW optimizer with a batch size of 16 and a learning rate of $1 e ^{-3}$, utilizing 2 A6000 GPUs.
\vspace{-6pt}
\subsubsection{Inference Settings}
The sampling steps $T$ is set to $50$. The guidance scale and image guidance scale are adjusted according to the specific tasks. More implementation details are provided in the supplementary material.
\vspace{-6pt}
\subsection{Text to Image}
\subsubsection{Qualitative Results}
We compare the text2ODI performance of proposed Omni$^2$ with other $3$ models trained on their respective datasets. The qualitative comparison results are presented in Fig. \ref{fig:comparison_generation}. Images generated by MVDiffusion \cite{mvdiffusion} are paded gray in the top and bottom since MVDiffusion can only generate images with $90^{\circ}$ vertical FoV. The images are rotated by $90^{\circ}$ to better visualize left-right consistency. As can be observed, ODIs generated by Text2Light \cite{text2light} exhibit poor left-right consistency and fail to capture detailed object specified in the text prompt, such as the missing “red building” in (a) and the “hot air balloon” in (b).
Despite the top-bottom view missing problem, MVDiffusion tends to generate repeating objects across different viewports since the ODIs are stitched from 8 viewports generated by SD \cite{sd} blocks, seperately, resulting in a loss of overall image coherence. ODIs generated by PanFusion \cite{panfusion} suffer from top-bottom blurriness and also exhibit insensitivity to specific details in the text prompt, \textit{e.g., }missing illuminated buildings in (c). In contrast, our model is capable of generating ODIs with clear $360^{\circ}\times 180^{\circ}$ FoV, while preserving fine details from the text prompt, demonstrating superior text2ODI performance for both indoor and outdoor ODI generation.
\vspace{-6pt}
\subsubsection{Quantitative Results}
Table \ref{tab:generation} presents the quantitative results of text2ODI task. We also LoRA-finetuned Stable Diffusion~\cite{sd} to provide more comparative data. We use Fréchet Auto-Encoder Distance (FAED) \cite{oh2022bips}, Fréchet Inception Distance (FID) \cite{fid}, Inception Score (IS) \cite{is} and Clip Score (CS) \cite{cs} to compare the quality of ODIs generated by Omni$^2$ with that by other methods. We also report inference time on a single A6000 for efficiency comparison. It can be observed that our model outperforms others in terms of all the evaluation metrics. Moreover, our method has a substantial advantage in inference time due to the attention mechanism using kv-cache.
\vspace{-3pt}
\subsubsection{User Study}
To better quantitize the performance of different methods, we collect 108 text prompts and recruit 20 volunteers to rate the generated ODIs from three perspectives: image quality, image-text consistency and omni-scene consistency. Experimental results presented in Table \ref{tab:userstudy} show that our proposed Omni$^2$ exhibits an overall superior performance, demonstrating its strong capability of ODI generation.
\vspace{-9pt}
\subsection{Multi-modal to Image}
% \subsubsection{Outpainting}
We compare the performance of our model with the state-of-the-art methods on the ODI outpainting task. For image-based outpainting, we report FID and IS scores for comparison. For text-guided outpainting, we add the CS metric to evaluate the correspondence between image and text. The results are presented in Table \ref{tab:outpainting}. Our model achieves state-of-the-art performance on ODI outpainting tasks, both with and without text prompts.
\begin{table}

\centering

\caption{Comparison with state-of-the-art methods on outpainting task.
%The cross-dataset evaluations of ``other datasets $\rightarrow$ KVQ".
}
\vspace{-2mm} 
\centering
\scalebox{0.8}{
% \resizebox{\linewidth}{!}
\begin{tabular}{lccccccc}
   \toprule
   {Task}&\multicolumn{3}{c}{Image Input}&\multicolumn{4}{c}{Text-Image Input}\\
   \cmidrule(lr){2-4} \cmidrule(lr){5-8} 
   Methods& FAED$\downarrow$& FID$\downarrow$&IS$\uparrow$& FAED$\downarrow$&FID$\downarrow$&IS$\uparrow$&CS$\uparrow$\\
     \midrule
   SIG-SS~\cite{SIG-SS}&1.08&66.67&5.04&N/A&N/A&N/A&N/A\\
OmniDreamer~\cite{omnidreamer}&1.83&73.80&5.15&N/A&N/A&N/A&N/A\\
   PanoDiffusion~\cite{PanoDiffusion}&1.54&127.30&4.19&N/A&N/A&N/A&N/A\\
   PanoDiff~\cite{panodiff}&2.21&61.03&6.30&1.22&45.54&{6.78}&0.8535\\
   \rowcolor{gray!20}
   Omni$^2$ (Ours)&\textbf{1.00}&\textbf{44.13}&\textbf{6.86}&\textbf{0.99}&\textbf{37.40}&\textbf{6.93}&\textbf{0.8620}\\
   \bottomrule
  \end{tabular}}
  \vspace{-1.5em}
  \label{tab:outpainting}
\end{table}

The qualitative results of image conditioned outpainting are presented in Fig. \ref{fig:comparison_outpaint}. We only provide the outpainting results from center viewport here, outpainting results from diverse input masks are presented in the supplementary. It should be noted that since we do not know the specific data split for these models, images in our test set could be in their training set. As can be seen in the figure, outpainting results generated by SIG-SS \cite{SIG-SS} and OmniDreamer \cite{omnidreamer} suffer from noticeable boundary inconsistencies, while PanoDiffusion \cite{PanoDiffusion} tends to produce images with evident artifacts. Although PanoDiff \cite{panodiff} demonstrates relatively better visual quality in terms of boundary consistency, the outpainting content appears less semantically reasonable. In contrast, our model significantly outperforms these baselines by generating semantically meaningful and visually coherent images with superior boundary consistency and enhanced overall visual fidelity. 

We also compare the performance of proposed Omni$^2$ with PanoDiff on text-guided ODI outpainting, qualitative results are shown in Fig. \ref{fig:comparison_outpaint1}. PanoDiff struggles to generate semantically plausible results, \textit{e.g. ,} grass in the sky in Fig. \ref{fig:comparison_outpaint1} (a). In addition, artifacts and meaningless objects appear in the generated ODIs. In contrast, our proposed Omni$^2$ is able to generate meaningful content that is consistent with the text prompt.

Due to the limited number of dedicated methods for ODI inpainting, semantic2image, and depth2image, the qualitative results of these tasks are presented in the supplementary material.
% \subsubsection{Inpainting}
% \input{tabs/inpainting}
% We compare the performance of our model with the state-of-the-art ODI image inpainting methods. For image-based inpainting, we report FID, sFID, and IS for comparison. For text-guided inpainting, we add the CS metric to evaluate the correspondence between image and text. The quantitative results are reported in Tab \ref{tab:inpainting}.
% \input{figs/comparison_inpaint}
\vspace{-9pt}
\subsection{Omnidirectional Image Editing}
Since there is no existing ODI editing method, we compare our method with existing 2D image editing methods in two aspects: directly applying existing editing methods on ODI, and editing the designated viewport separately after viewport splitting.

We first compare our method with directly applying 2D editing on ODIs. The results are shown in Fig. \ref{fig:comparison_editing}. As illustrated, HQ-Edit \cite{hqedit} generates images that are completely unrelated to the source image, while InstructPix2Pix \cite{instructpix2pix} fails to understand the instructions and produces incorrect images, such as the wall being mistakenly painted yellow in Fig. \ref{fig:comparison_editing}(b). Although MagicBrush \cite{magicbrush} seems to partially understand some of the editing instructions and makes adjustments to the source image, it struggles with understanding the direction of the edits, which is critical for immersive VR editing. Furthermore, all these methods fail to truly understand the object within the image, as can be clearly seen in fig. \ref{fig:comparison_editing}(a), possibly due to the fact that 2D editing models are not trained to understand the warped ERP format.

We also compare our method with applying 2D editing method on designated views. The results for the edited viewport are presented in Fig. \ref{fig:comparison_editing1}. We split the view such that the targeted object is centered in the viewport image. As can be observed, both InstructPix2Pix and HQ-Edit fail to comprehend the instructions while MagicBrush introduces noticeable artifacts, as highlighted in the yellow box. Additionally, the added object lacks the level of detail produced by our method. More importantly, this split-and-edit approach is not applicable to real-world scenarios, as it overlooks the global coherence of the entire 360° image and the complex interactions between different regions of the scene.

The experimental results clearly demonstrate the ineffectiveness of applying 2D editing methods on ODIs, which underscores the importance of developing methods tailored to the unique demands of omnidirectional image editing.
\vspace{-6pt}
\subsection{Ablation Study}
In this section, we conduct ablation studies to validate the utility of the core modules in Omni$^2$.
\subsubsection{Bidirectional Attention Mechanism}
We adopt bidirectional attention within the viewport sequence to maintain viewport consistency. Fig. \ref{fig:abl}-Top shows the text2image result with causal attention. As can be seen, great distortion occurs, and there are clear boundries for overlapping viewports. The result demonstrate the effectiveness of the proposed viewport-based bidirectional attention.
\subsubsection{Loss Function}
\begin{figure}[t]\centering
\includegraphics[width=0.95\linewidth]{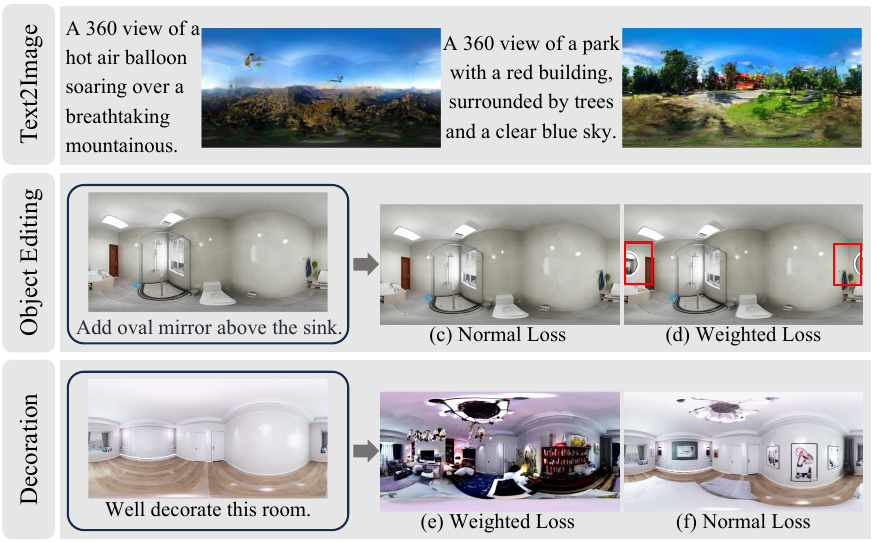}
\vspace{-1em}
\captionof{figure}{Visual results of ablation studies. \textbf{Top}: ablation on viewport-based bidirectional attention; \textbf{Middle}: ablation on weighed loss for object-level editing tasks; \textbf{Bottom}: ablation on normal loss for decoration task.
}
\label{fig:abl}
\vspace{-1.5em}
\end{figure}

We modify the loss function so that the model learns to modify dedicated regions within the input image while keeping other parts unchanged. For the decoration task, we use the origin loss for better results. Ablation studies are conducted to demonstrate the effectiveness of the modified loss function, with results presented in Fig. \ref{fig:abl}-Middle and -Bottom. For object-level image editing, the model tends to simply copying the input as output without weighted loss. For the decoration task, since modifications are made on the whole image, applying weighted loss leads to heavy distortion in generated image.
\subsubsection{LoRA Rank}
The LoRA rank $r$ is set to $16$ in the paper, we also compare the influence of different rank and report the quantitative results of text2image task in Table \ref{tab:abl}. As shown in the table, using $r=16$ generally performs better than using $r=8$ and $r=32$, which further validates the effect of LoRA finetuning.
%\begin{figure*}[t]\centering
%\vspace{-0.8em}
%\includegraphics[width=0.98\linewidth]{figs/figs/fig1.pdf}
%\vspace{-0.6em}
%\caption{
%}
%\label{fig:1_frontpage}
%\vspace{-1em}
%\end{figure*}

%\vspace{-1em}
% \vspace{-6pt}
\begin{table}

\centering

\caption{Ablation study on LoRA rank, with T2I performance results reported.}
\label{tab:abl}
\vspace{-4mm} 
\centering
% \resizebox{\linewidth}{!}
\scalebox{0.8}{
\begin{tabular}{ccccc}
   \toprule
   \textit{LoRA rank $r$}& FID$\downarrow$&IS$\uparrow$&CS$\uparrow$\\
     \midrule
   8&48.68&6.96&0.8873\\
   % 16& & & \\
   32&47.63&7.34&0.8853\\
   \rowcolor{gray!20}
   16 (Ours)&\textbf{47.32}&\textbf{7.62}&\textbf{0.8887}\\
   \bottomrule
  \end{tabular}}
\vspace{-1.5em}
\end{table}
  % \vspace{-1.5em}

\vspace{-3pt}
\section{Conclusion}
In this paper, we aim to unify omnidirectional image generation and editing tasks.
Specifically, we first construct Any2Omni, the first comprehensive dataset containing 
60,000+ data encompassing various ODI generation and editing tasks. Any2Omni contains the first comprehensive multi-task ODI generation subset with diverse input conditions and the first ODI editing subset featuring both object-level and scene-level editing tasks. 
Based on the dataset, we propose Omni$^2$, an omni model for omnidirectional image generation and editing via a Transformer architecture with viewport-based bidirectional attention mechanism, which is able to process multi-modal input conditions and generate high-quality ODIs across various tasks.
Extensive experiments demonstrate that our proposed model achieves state-of-the-art performance on various ODI generation tasks and exhibits strong potential for ODI editing tasks.
%%
%% The next two lines define the bibliography style to be used, and
%% the bibliography file.
\section{Acknowledgments}
This work was supported in part by the National Natural Science Foundation of China under Grants 62401365, 62225112, 62271312, 62132006, U24A20220, and in part by the China Postdoctoral Science Foundation under Grant Number BX20250411, 2025M773473, and in part by STCSM under Grant 22DZ2229005.
\bibliographystyle{ACM-Reference-Format}
\balance
\bibliography{sample-base}
\newpage
\appendix
\section{More Details of Any2Omni Construction}\label{supp1}
% \subsection{Dataset composition}
%\begin{figure*}[t]\centering
%\vspace{-0.8em}
%\includegraphics[width=0.98\linewidth]{figs/figs/fig1.pdf}
%\vspace{-0.6em}
%\caption{
%}
%\label{fig:1_frontpage}
%\vspace{-1em}
%\end{figure*}

%\vspace{-1em}
\begin{figure}[t]\centering
\includegraphics[width=1\linewidth]{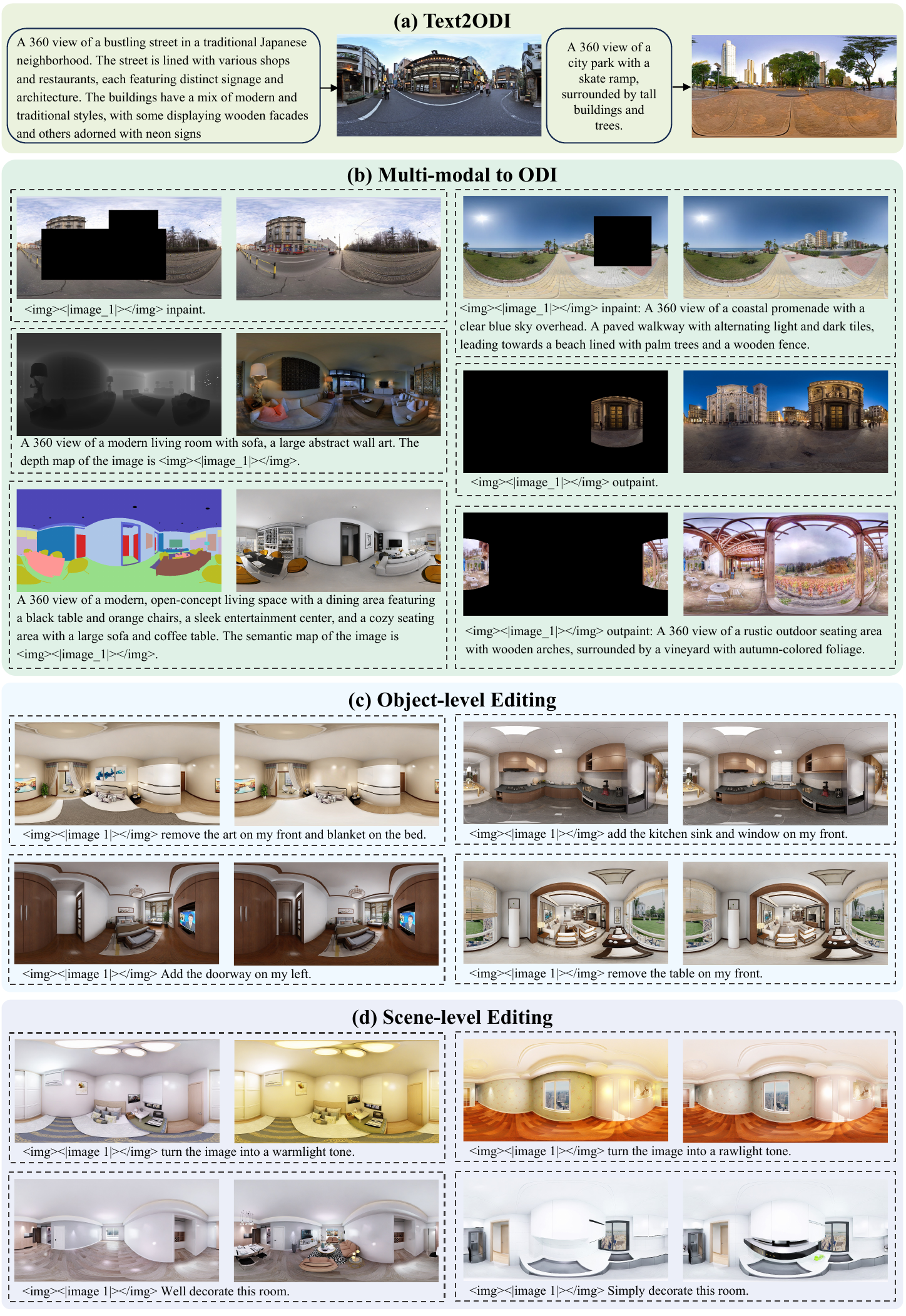}
\vspace{-2.5em}
\captionof{figure}{Examples of Any2Omni dataset. The input of all tasks are organized into an interleaved image-text sequence format.
}
\label{fig:dataset_overview}
\vspace{-1em}
\end{figure}
% \vspace{-6pt}
Fig. \ref{fig:dataset_overview} presents some examples of Any2Omni dataset. Detailed composition and task description of Any2Omni is presented below.
\subsection{Text to Image}
Input for this subset of data is text only. Although prior text2ODI works utilize captioned ODI datasets for training \cite{mvdiffusion, panfusion}, some are captioned only for individual views rather than the entire ODI \cite{mvdiffusion}, and there is no standardized annotation format for full ODIs \cite{panfusion, diffpano}. We aim to construct a comprehensive text-to-ODI dataset with both short and detailed text descriptions for diverse ODI generation.

Leveraging the powerful multi-modal understanding capabilities of vision-language models (VLMs), we employ InternVL2-5 \cite{internvl} to generate text descriptions for 20,000 ODIs from the SUN360 dataset \cite{sun360}. SUN360 encompasses a diverse range of indoor and outdoor scenes, making it well-suited for text-to-ODI training both in terms of scene diversity and dataset scale.
We adopt a dual-strategy approach for generating diverse text descriptions: for half of the images, we input length-unconstrained prompts to generate detailed text descriptions, while for the other half, we restrict prompts to concise the output text of no more than 20 words. This approach ensures that our dataset supports both generating high-quality ODIs from brief prompts and producing highly-detailed text-aligned ODIs based on fine-grained text descriptions.
\subsection{Multi-modal to Image}
We integrate existing ODI generation tasks and propose new ones utilizing existing ODI data to create diverse multi-modal to ODI training examples. Examples of this subset is presented in Fig. \ref{fig:dataset_overview}(b). Trained on these data, our Omni$^2$ model is capable of supporting a wide range of task types with multi-modal inputs, enabling more versatile and comprehensive ODI generation capability.

\subsubsection{Inpainting}
We select image from SUN360 \cite{sun360} dataset as the ground truth and generate masks with random shapes including rectangular, irregular and single-view masks, to create the inpainting dataset. The inpainting task is further categorized into text-guided inpainting, \textit{i.e.}, the masked image and GT description generated by Internvl2-5 are used as input, and non-text-guided inpainting, where only the masked image is used.
\subsubsection{Outpainting}
We select images from the SUN360 dataset and extract partial viewpoints as input for outpainting. Similar to inpainting, the outpainting task is also divided into text-guided and non-text-guided categories, depending on whether a description is provided alongside the input image.
\subsubsection{Depth to Image}
ODIs are highly relevant to depth information, and some existing works have explored depth estimation for ODIs \cite{pano3d, li2021panodepth, junayed2022himode}. However, few works have been done to generate ODIs with the aid of $360^{\circ}$depth information. To bridge this gap, we introduce a Depth2Image task for ODI and construct a dedicated subset using existing ODI depth estimation datasets. Specifically, we select images and their corresponding depth maps from the Pano3D dataset \cite{pano3d} and generate scene descriptions using Internvl2-5.
\subsubsection{Semantic to Image}
Structured3D dataset \cite{structured3d} offers a diverse collection of indoor ODIs and their corresponding semantic segmentation maps. Similarly to 2D images, we define a semantic2image task for omnidirectional images. We utilize Internvl2-5 to generate descriptions for the ODIs and incorporate both these descriptions and the semantic segmentation maps as inputs to train the model that produces semantically coherent omnidirectional images.
\subsection{Object-level ODI Editing}
We utilize the pipeline proposed in the main paper to construct the first object-level ODI editing datasets, including single-object removal/addition and multi-object removal/addition. Notably, the simple pipeline we propose supports the generation of large-scale object-level ODI editing datasets. In this work, to ensure balanced training data for each task, we generate a total of over 12,000 editing data entries across thousands of scenes. Given the richness of semantic information in indoor ODI environments, the object-level ODI editing can be further categorized into: single object removal/add, multi-object removal/add in one view and multi-object removal/add across different views. Some examples are presented in Fig. \ref{fig:dataset_overview}(c).
\subsection{Scene-level ODI Editing}
Apart from using synthetic data pipeline for generating object-level editing, we propose two insightful scene-level indoor editing tasks utilizing data in Structured3D dataset \cite{structured3d} as shown in Fig. \ref{fig:dataset_overview}(d). Structured3D is a large photo-realistic dataset with high-quality rendered images of various types, thus can be well applied for image editing tasks.
\subsubsection{Light-Modify}
Indoor lighting modification plays a crucial role in virtual reality experiences, and has broad application in real life. Structured3D dataset utilizes industry-leading rendering engines to simulate photo-realistic indoor scenes under different lighting conditions. We select data from this dataset and generate corresponding instructions to build a dataset for indoor light-modify tasks.
\subsubsection{Decoration}
Structured3D generates different configurations (full, simple and empty) of the same room by removing some or all furniture. Building upon this dataset, we further define an interesting and meaningful task, termed indoor decoration. We utilize the data from Structured3D under different configurations and generate instructions to construct a dataset for this task.

\section{More Details of Experiment Settings}
\subsection{Training and Inference Details}
\begin{table}

\centering

\caption{Inference settings for different tasks.}
\label{tab:guidance_scale}
\vspace{-4mm} 
\centering
% \resizebox{\linewidth}{!}
\scalebox{0.8}{
\begin{tabular}{lcc}
   \toprule
   Tasks& Guidance Scale&Image Guidance Scale\\
     \midrule
   Text2ODI&2.5&N/A\\
   % \hdashline
   Inpainting&2.5&1.8\\
   Outpainting&2.5&1.8\\
   Depth2Image&2.0&1.8\\
   Semantic2Image&2.0&1.8\\
   \hdashline
   Object-level Editing&3.0&1.8\\
   Light-modify&3.0&1.8\\
   Decoration&3.5&1.8\\
   \bottomrule
  \end{tabular}}
\vspace{-1em}
\end{table}
  % \vspace{-1.5em}

Omni$^2$ is able to process interleaved texts and images as input conditions. This is achieved by \textbf{joint training} with data from multiple tasks.
We train the model with a resolution of $256\times 256$ and FoV of $100^{\circ}$ for perspective images for 10k steps, with a batch size of $16$, utilizing two A6000 GPUs.
Guidance scale is set as the strength of the text guidance, the larger the guidance, the more similar the generated image will be to the prompt. Image guidance scale indicates the guidance strength of the input image, where larger image guidance leads to generating images closer to the input image condition. During inference, guidance scale and image guidance scale are set based on different tasks to achieve better visual results, as presented in Table \ref{tab:guidance_scale}.

\section{More Details of Comparison Experiment}
\subsection{Text2ODI}
\subsubsection{Baseline Models}
\begin{itemize}
    \item Text2Light~\cite{text2light} adopts a two-stage approach that first generates a low-resolution ODI based on the input text, and then expands it to ultra-high resolution. The model provides two sets of checkpoints for indoor and outdoor ODI generation, during inference, we adopt these two checkpoints based on the input prompt.
    \item MVDiffusion~\cite{mvdiffusion} generates 8 viewports using pretrained SD \cite{sd} blocks seperately and fine-tunes the inserted CAA block for multi-view consistency. We adopt the pretrained weights for comparison purposes. During training, each viewport is assigned a viewport-specific prompt.However, during infernece, the model lacks the ability to process the prompt for the entire ODI; instead, it simply copies the prompt for each viewport, leading to noticeable object repetition across different views. Since the generated image lacks top and bottom views, the quantitative results may not be completely reliable and are presented in the main paper for reference only.
    \item PanFusion~\cite{panfusion} is a text-to-ODI model designed to reduce distortion caused by projecting perspective images onto an ODI canvas, while also offering global layout guidance. However, the generated images still suffer from quality issues, such as repeated objects appearing across different views, which significantly affects the visual coherence and overall aesthetic quality. Additionally, noticeable blurriness is observed at the top and bottom regions of the generated ODIs, as discussed in the main paper.
\end{itemize}
\subsubsection{More baseline comparison}
We further adapt MVDiffusion \cite{mvdiffusion} to generate six viewports as our model and retrain it on our database and LoRA-finetuned Stable Diffusion \cite{sd} and Flux \cite{flux2024} on Any2Omni T2I dataset for more baseline comparison. The results are reported in Table \ref{tab:more_baseline}. As reported in the table, MVDiffusion shows great overfitting after retraining. Our model attains an overall state-of-the-art performance.
\begin{table}

\centering

\caption{Comparison with more baselines on text to omnidirectional image task.}
\vspace{-3mm} 
\centering
\resizebox{0.8\linewidth}{!}{
\begin{tabular}{lcccc}
   \toprule
   Methods& FAED$\downarrow$& FID$\downarrow$&IS$\uparrow$&CS$\uparrow$\\
     \midrule
   MVDiffusion (retrained)~\cite{mvdiffusion}&3.11&69.21&5.58&0.564 \\
   SD+LoRA~\cite{panfusion}&2.30&57.97&7.41&0.854\\
   Flux+LoRA~\cite{sd}&2.53&53.87&\textbf{7.87}&0.873\\
   \rowcolor{gray!20}
   Omni$^2$ (Ours)&\textbf{2.25}&\textbf{47.32}&{7.62}&\textbf{0.888}\\
   \bottomrule
  \end{tabular}}
  % \vspace{-2em}
  \label{tab:more_baseline}
\end{table}
  % \vspace{-1.5em}

\subsubsection{User Study}
\begin{figure*}[t]\centering
\includegraphics[width=1\linewidth]{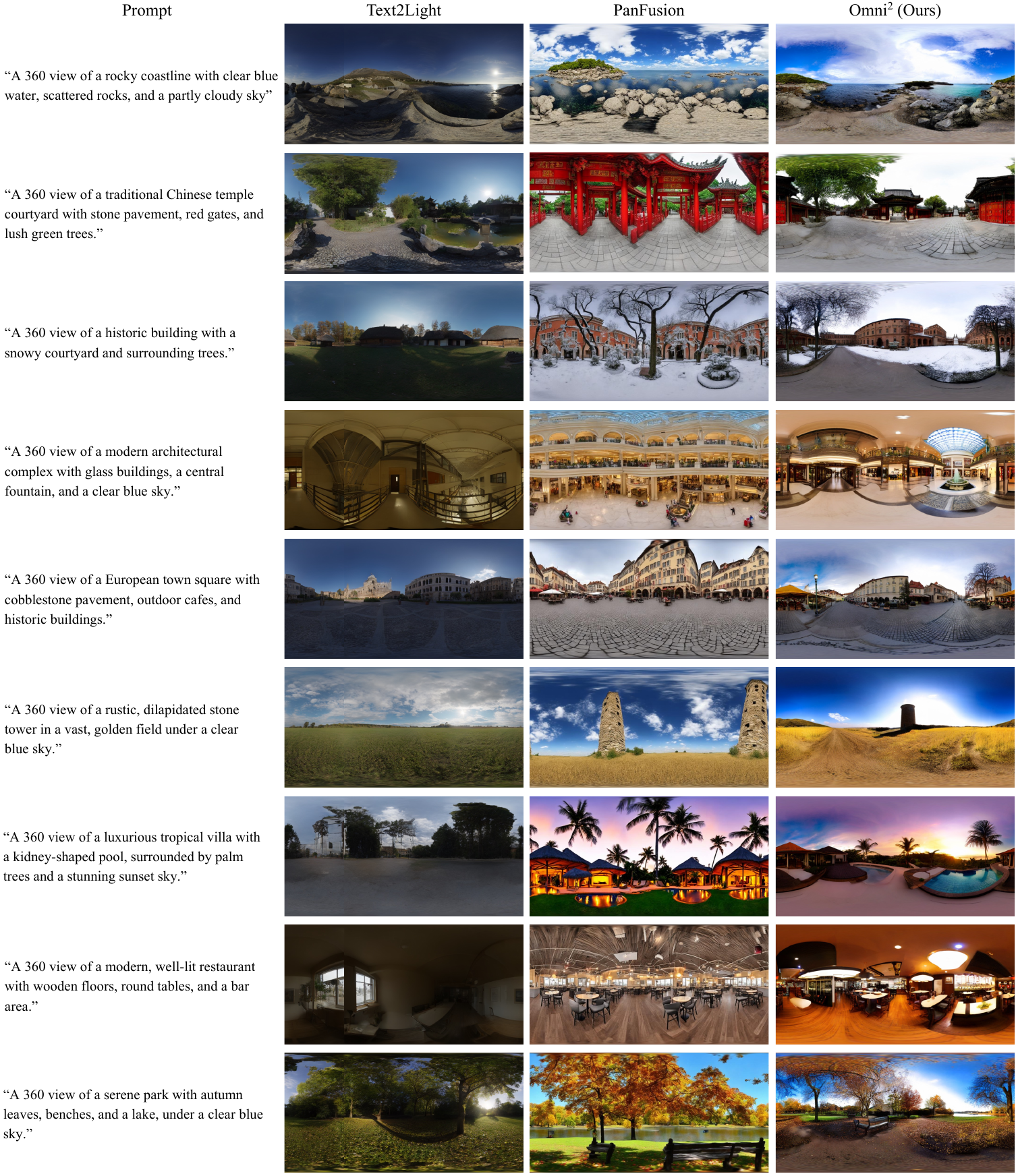}
\vspace{-2em}
\captionof{figure}{Examples of generated ODIs for user study.
}
\label{fig:userstudy1}
\vspace{-1.5em}
\end{figure*}

\begin{figure*}[t]\centering
\includegraphics[width=1\linewidth]{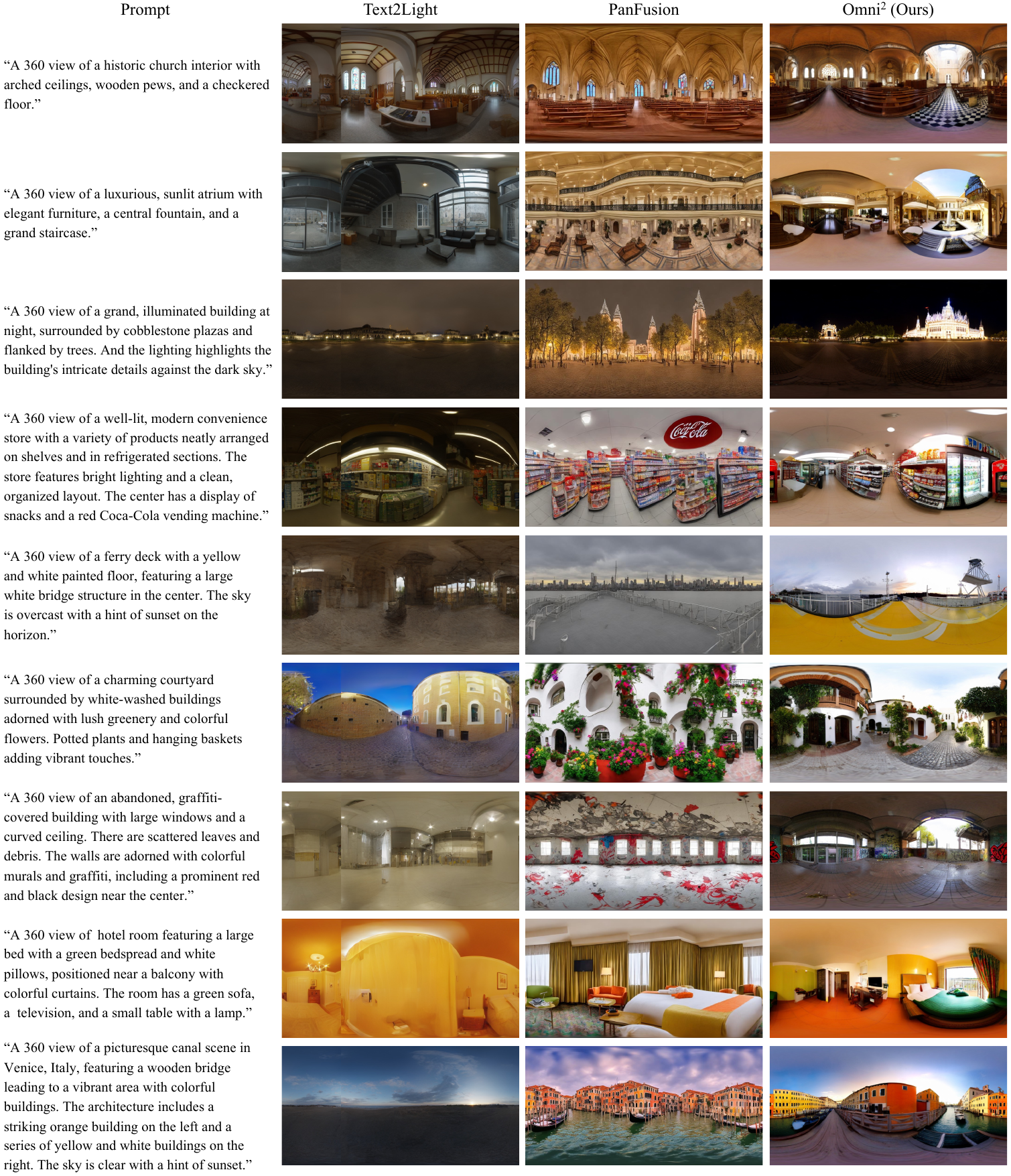}
\vspace{-2em}
\captionof{figure}{Examples of generated ODIs for user study.
}
\label{fig:userstudy2}
\vspace{-1.5em}
\end{figure*}

User study is conducted to a human preference perspective to provide additional insights for comparing text2ODI methods. MVDiffusion is excluded from the comparison as it is unable to generate full Field-of-View (FoV) ODIs.
We select 108 prompts, covering both indoor and outdoor scenarios for generating ODIs. Examples of the prompts and generated ODIs are presented in Fig. \ref{fig:userstudy1} and Fig. \ref{fig:userstudy2} to serve as more T2ODI comparison results. The images are rotated by $90^{\circ}$ to show left-right consistency.

Participants are instructed to rate the generated ODIs on a scale from 1 to 5, in increments of 1, from the following three perspectives:
\paragraph{Image Quality:}
The overall quality of the generated ODI, including both low-level aspects (\textit{e.g., }color, brightness, \textit{etc.}) and high-level attributes (\textit{e.g., }authenticity, aesthetic appeal and scene coherence). (1) Poor: The ODI contains severe distortions, with noticeable noise, unnatural colors, and numerous artifacts, all of which significantly impact viewing comfort and visual coherence. (2) Bad: The overall image quality is low, with visible noise and artifacts. While distortions are present, they are less disruptive compared to the “Poor” level. (3) Fair: There are slight distortions within the generated ODI, the low-level quality is acceptable but there are artifacts and unrealistic scenes that affect the aesthetic quality. (4) Good: The overall image quality is high, with minimal distortions and natural colors. However, the scene may still contain elements that appear somewhat unrealistic, slightly affecting authenticity. (5) Excellent: No distortions occur in the generaed ODI, the color is natural and colorful, the image is smooth and the scene is coherent and closely resembles real-world environments.
\paragraph{Image-Text Consistency: }
The degree to which the generated image aligns with the content of the text prompt. (1) Poor: The scene is totally irrelevant to the text descriptions. (2) Bad: The scene is of low relavance to the text descriptions, while the general background may loosely match, the specific details mentioned in the prompt are missing or incorrect. (3) Fair: The generated image captures some key elements of the text description, but several important details are either inaccurate, missing, or only partially reflected. (4) Good: Most of the content described in the prompt is accurately represented in the generated image, though a few minor details may be inconsistent or underrepresented. (5) Excellent: The generated image aligns very closely with the text prompt, with both global structure and fine-grained details accurately depicted.
\paragraph{Omni-Scene Consistency}
The degree of consistency across the scene, including left-right alignment and overall scene coherence. (1) Poor: The left and right regions are completely inconsistent, depicting different or unrelated scene content. (2) Bad: The left and right views are misaligned, though they attempt to depict similar content. Clear discontinuities are present. (3) Fair: Left-right consistency is generally acceptable, but inconsistencies remain across adjacent viewports, affecting the perception of a coherent 360° scene. (4) Good: The overall scene consistency is ok with only minor mismatches or discontinuities in certain regions. (5) The scene is fully consistent across all directions, with seamless transitions between viewports and no perceptible misalignments.

The mean opinion scores (MOS) for each model, averaged across all participants, are reported in the main paper for each evaluation criterion.
%\begin{figure*}[t]\centering
%\vspace{-0.8em}
%\includegraphics[width=0.98\linewidth]{figs/figs/fig1.pdf}
%\vspace{-0.6em}
%\caption{
%}
%\label{fig:1_frontpage}
%\vspace{-1em}
%\end{figure*}

%\vspace{-1em}
% \vspace{-12pt}
%\begin{figure*}[t]\centering
%\vspace{-0.8em}
%\includegraphics[width=0.98\linewidth]{figs/figs/fig1.pdf}
%\vspace{-0.6em}
%\caption{
%}
%\label{fig:1_frontpage}
%\vspace{-1em}
%\end{figure*}

%\vspace{-1em}
% \vspace{-12pt}
\subsection{Outpainting}
\begin{figure*}[t]\centering
\includegraphics[width=1\linewidth]{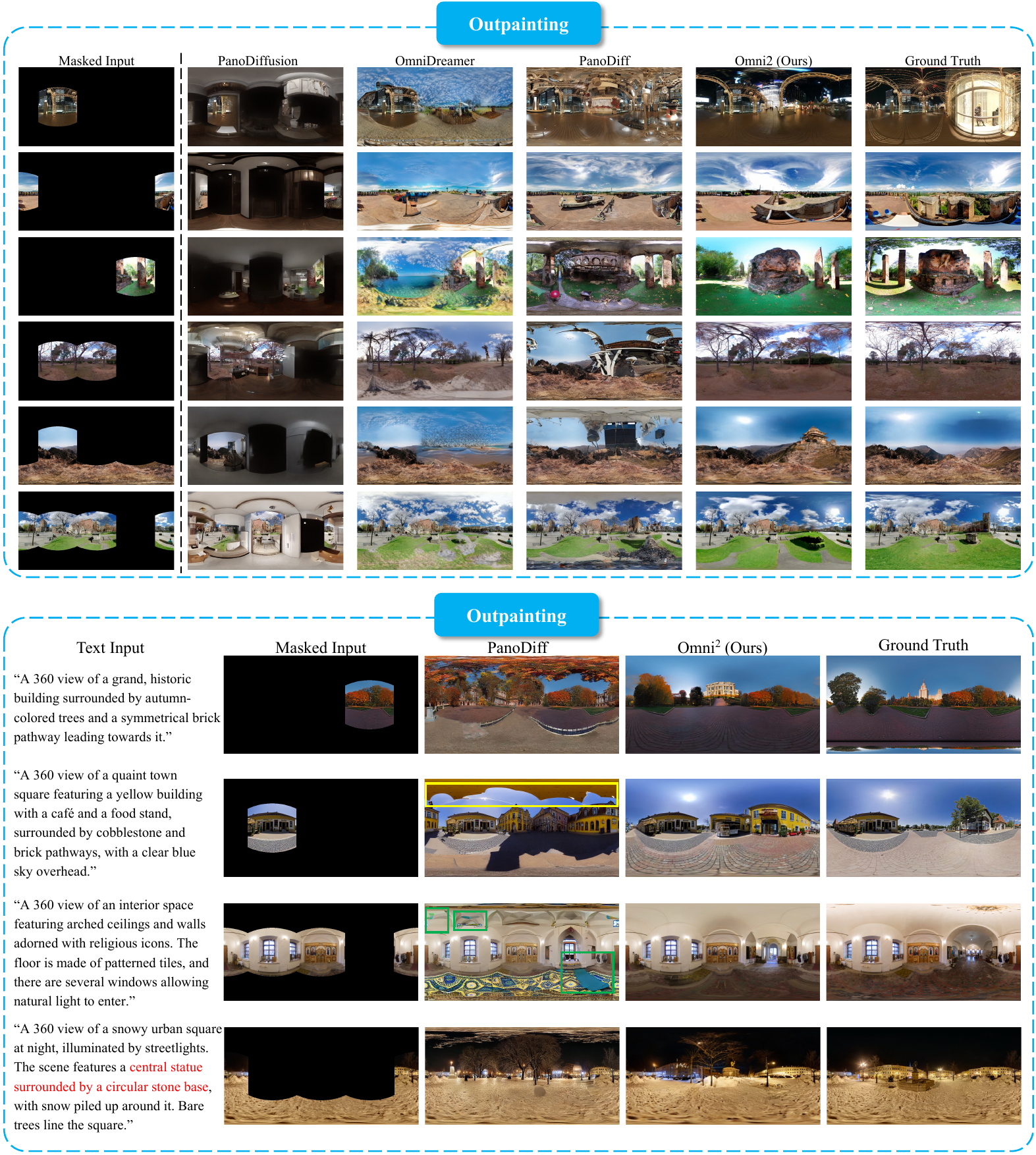}
\vspace{-2.5em}
\captionof{figure}{More comparison results of outpainting. Top: outpainting comparison between  PanoDiffusion \cite{PanoDiffusion}, OmniDreamer \cite{omnidreamer}, PanoDiff \cite{panodiff}, and ours from various masked input images. Bottom: Outpainting comparison between PanoDiff and ours from various masked input images.  We highlight the \textcolor{yellow}{semantic inconsistencies} and \textcolor{myspringgreen}{folating artifact} with corresponding color boxes. Key objects that are missing in PanoDiff but present in our method are highlighted in  \textcolor{red}{red} in the prompts.
}
\label{fig:outpaint_supp_all}
\vspace{-1.5em}
\end{figure*}

We provide more outpainting results from diverse viewport input masks to better show the performance of our proposed Omni$^2$. The results without text guidance are provided in Fig. \ref{fig:outpaint_supp_all}-Top. Our model shows state-of-the-art performance on image-input outpainting tasks.

Fig. \ref{fig:outpaint_supp_all}-Bottom presents the results of proposed Omni$^2$ and PanoDiff \cite{panodiff} from text-image input. PanoDiff is an ODI outpainting model trained on SUN360 dataset, which may have overlap with our testing set. As can be seen from the figure, outpainting results of PanoDiff suffer from semantic inconsistencies within the scenes and noticeable floating artifacts especially in the top and bottom viewports. Moreover, the model demonstrates limited sensitivity to the detailed content described in the text prompt.
%\begin{figure*}[t]\centering
%\vspace{-0.8em}
%\includegraphics[width=0.98\linewidth]{figs/figs/fig1.pdf}
%\vspace{-0.6em}
%\caption{
%}
%\label{fig:1_frontpage}
%\vspace{-1em}
%\end{figure*}

%\vspace{-1em}
% \vspace{-6pt}
%\begin{figure*}[t]\centering
%\vspace{-0.8em}
%\includegraphics[width=0.98\linewidth]{figs/figs/fig1.pdf}
%\vspace{-0.6em}
%\caption{
%}
%\label{fig:1_frontpage}
%\vspace{-1em}
%\end{figure*}

%\vspace{-1em}
\begin{figure*}[t]\centering
\includegraphics[width=1\linewidth]{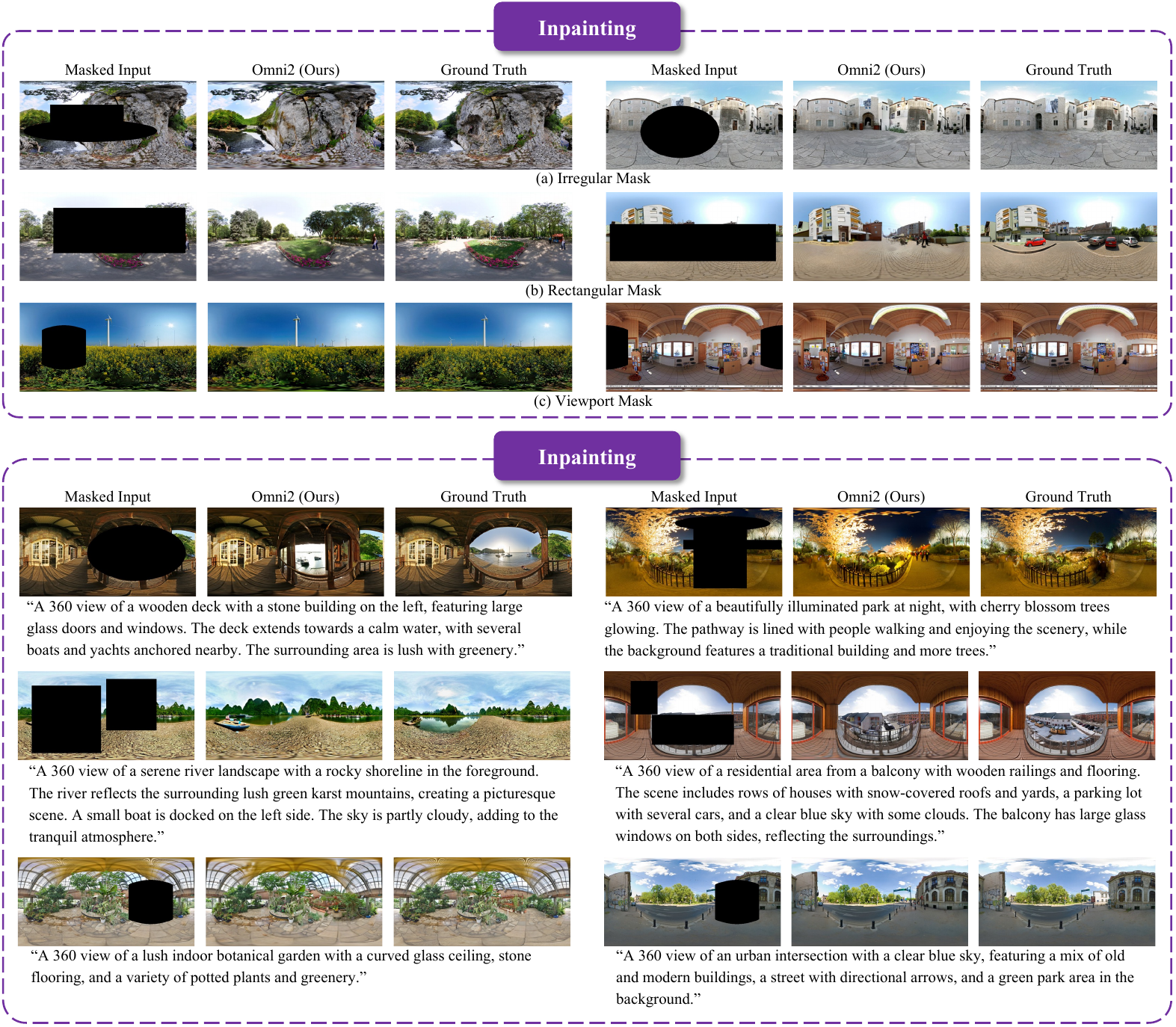}
\vspace{-2.5em}
\captionof{figure}{Qualitative results of ODI inpainting with Omni$^2$. Top: ODI inpainting using a single masked image as input. Bottom: ODI inpainting guided by both a masked image and text descriptions.
}
\label{fig:inpaint_supp}
\vspace{-1.5em}
\end{figure*}
% \vspace{-6pt}

\section{More visualization results of Omni$^2$}
\subsection{Inpainting}
As there are limited ODI inpaiting methods, we only present qualitative results in Fig. \ref{fig:inpaint_supp} for visualization. 
% As can be seen from the figure, our proposed Omni$^2$ is capable of scene consistent inpainting results for various shapes of input masks. Furthermore, the generated ODIs exhibit strong alignment with the optional text guidance.
Fig. \ref{fig:inpaint_supp}-Top presents ODI inpaint with a single masked image as input, different mask shapes are adopted. Our Omni$^2$ is capable of generating plausible content within the masked regions while maintaining overall scene consistency. The results of ODI inpainting with both text guidance and masked image are presented in Fig. \ref{fig:inpaint_supp}-Bottom presents inpainting results guided by both the masked image and a text prompt. As illustrated, Omni$^2$ excels at incorporating textual guidance to produce seamless, text-aligned ODIs.

\subsection{Semantic2Image}
The input conditions for this task are semantic maps and the corresponding scene descriptions. Our proposed Omni$^2$ demonstrates superior performance on these newly introduced tasks, generating images that exhibit high alignment with the text descriptions based on the semantic maps. The qualitative results are presented in Fig. \ref{fig:supp_gen}-Top.

\subsection{Depth2Image}
\begin{figure*}[t]\centering
\includegraphics[width=1\linewidth]{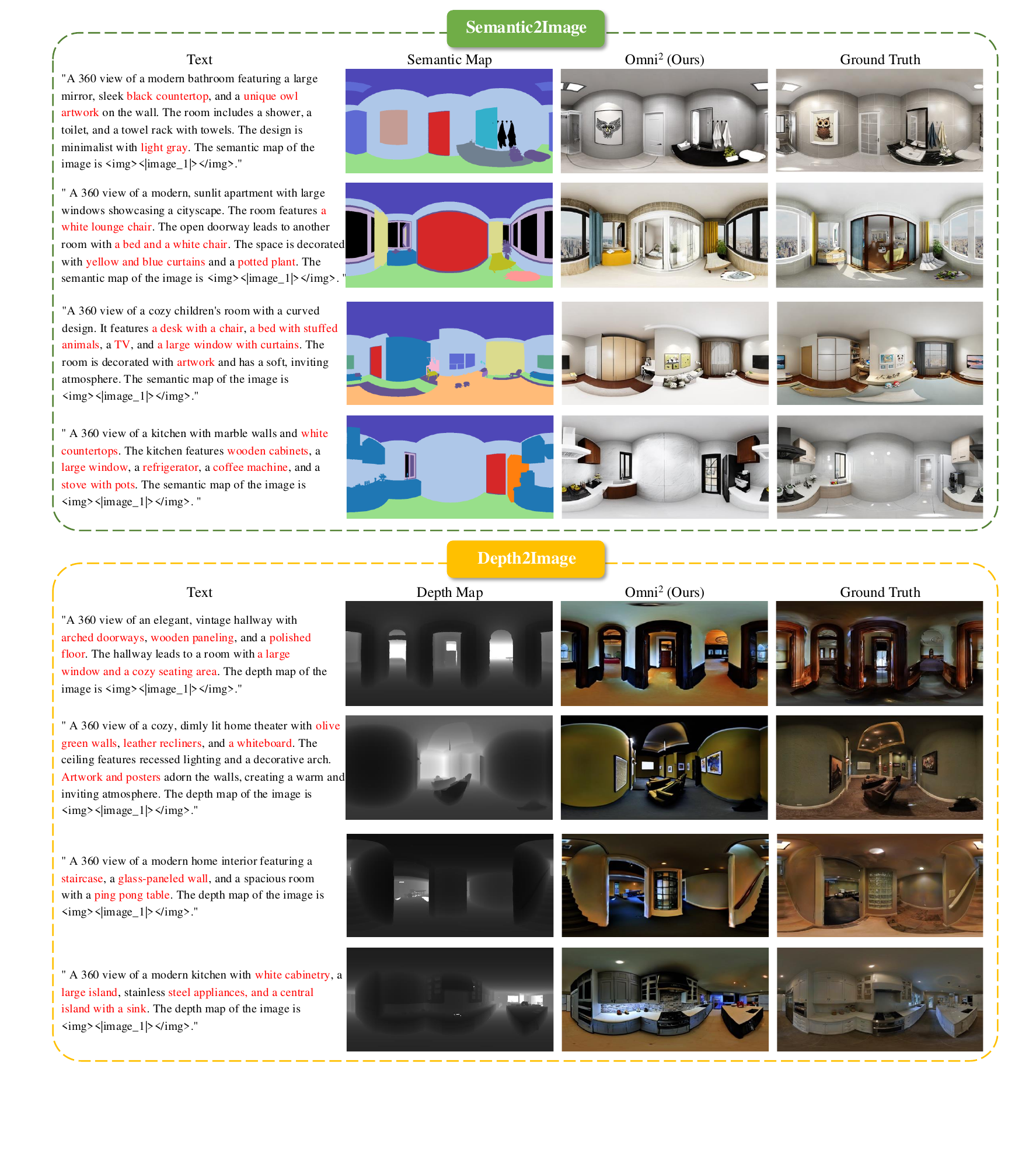}
\vspace{-2em}
\captionof{figure}{Qualitative results of Semantic2Image (top) and Depth2Image (bottom). Key words in the text prompt are marked in \textcolor{red}{red}.}
\label{fig:supp_gen}
\vspace{-1.5em}
\end{figure*}

The input conditions for this task are depth maps and the corresponding scene descriptions. Qualitative results of this task is presented in Fig. \ref{fig:supp_gen}-Bottom. The performance on this task is currently limited by the quality of the training dataset, which can be enhanced in future work.
\subsection{ODI Editing}
\begin{figure*}[t]\centering
\includegraphics[width=1\linewidth]{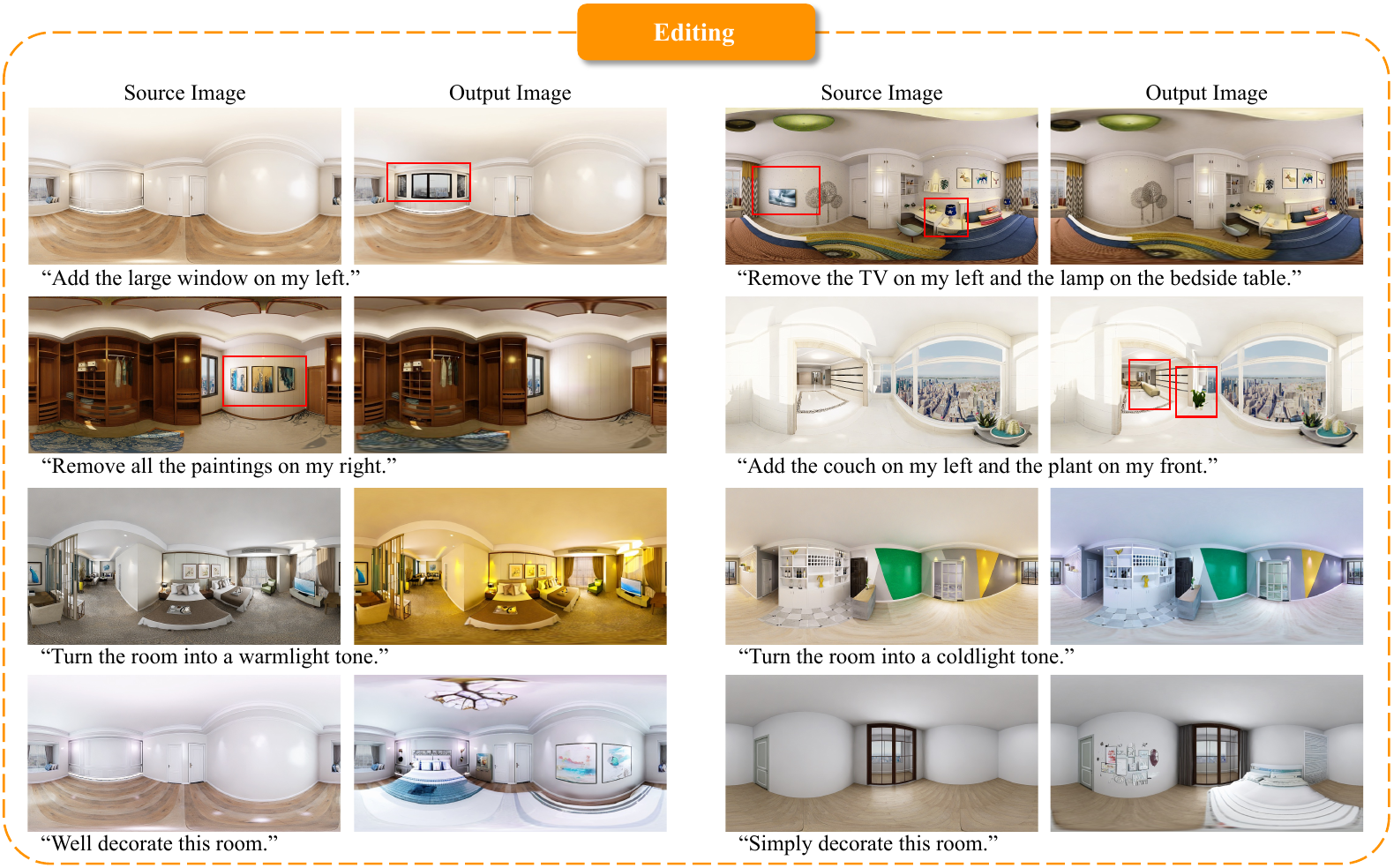}
\vspace{-2em}
\captionof{figure}{Qualitative results of object-level (top two rows) and scene-level (bottom two rows) editing tasks.}
\label{fig:supp_edit}
\vspace{-1.5em}
\end{figure*}

As discussed in the main paper, we are the first to explore editing tasks specifically for omnidirectional images, where existing 2D image editing methods are not directly applicable. In this section, we provide more qualitative of Omni$^2$ on ODI editing tasks, including both object-level editing and scene-level editing. The results are shown in Fig. \ref{fig:supp_edit}. Omni$^2$ demonstrate strong capabilities across these proposed editing tasks.
%\begin{figure*}[t]\centering
%\vspace{-0.8em}
%\includegraphics[width=0.98\linewidth]{figs/figs/fig1.pdf}
%\vspace{-0.6em}
%\caption{
%}
%\label{fig:1_frontpage}
%\vspace{-1em}
%\end{figure*}

%\vspace{-1em}
% \vspace{-12pt}
%\begin{figure*}[t]\centering
%\vspace{-0.8em}
%\includegraphics[width=0.98\linewidth]{figs/figs/fig1.pdf}
%\vspace{-0.6em}
%\caption{
%}
%\label{fig:1_frontpage}
%\vspace{-1em}
%\end{figure*}

%\vspace{-1em}
% \vspace{-12pt}

\end{document}